\documentclass{article}

\usepackage{arxiv}

\usepackage[utf8]{inputenc} % allow utf-8 input
\usepackage[T1]{fontenc}    % use 8-bit T1 fonts
\usepackage{hyperref}       % hyperlinks
\usepackage{url}            % simple URL typesetting
\usepackage{booktabs}       % professional-quality tables
\usepackage{amsfonts}       % blackboard math symbols
\usepackage{nicefrac}       % compact symbols for 1/2, etc.
\usepackage{microtype}      % microtypography
\usepackage{lipsum}
\usepackage{graphicx}
\graphicspath{ {./images/} }

\makeatletter
\DeclareRobustCommand\onedot{\futurelet\@let@token\@onedot}
\def\@onedot{\ifx\@let@token.\else.\null\fi\xspace}

\usepackage{enumitem}
\usepackage{upgreek}
\usepackage{bm}
\usepackage{wrapfig}
\usepackage{graphicx}
\usepackage{amsmath}
\usepackage{multirow}
\usepackage{mathtools}
\usepackage{floatrow}
\usepackage{paracol}
\usepackage{makecell}
\usepackage{csquotes}
\usepackage{comment}
\usepackage{dsfont}
\usepackage{xfrac}
\usepackage{subcaption}
\usepackage[noblocks]{authblk}

\usepackage{algorithm}
\usepackage[noend]{algorithmic}

\usepackage{amsthm}
\newtheorem*{definition}{Definition}

 \usepackage{array,multirow}
 \usepackage{float}
 \usepackage{rotating}
 \usepackage{booktabs}

\def\e{\mathbf{e}}
\def\x{\mathbf{x}}
\def\u{\mathbf{u}}
\def\w{\mathbf{W}}

\title{Brain-inspired Computational Intelligence \\ via Predictive Coding}
\author{% 
    \textbf{Tommaso Salvatori}$^{1,2}$ \ \ \  
    \textbf{Ankur Mali}$^{3}$ \ \ \  
    \textbf{Christopher L. Buckley}$^{1,4}$ \ \ \
    \textbf{Thomas Lukasiewicz}$^{2,5}$, \newline
    \textbf{Rajesh P. N. Rao}$^{6}$ \ \ \  
    \textbf{Karl Friston}$^{1,7}$ \ \ \
    \textbf{Alexander Ororbia}$^{8}$ \\
    $^1$\,VERSES AI Research Lab Los Angeles, California, USA\\
    $^2$\,Institute of Logic and Computation, Vienna University of Technology, Austria\\
    $^3$\,University of South Florida 
  Tampa, FL 33620, USA  \\
    $^4$\,Sussex AI Group, Department of Informatics, University of Sussex, Brighton, UK\\
    $^5$\,Department of Computer Science, University of Oxford, UK  \\
    $^6$\,Paul G. Allen School of Computer Science and Engineering, University of Washington 
  Seattle, Washington, USA  \\
    $^7$\,The Wellcome Centre for Human Neuroimaging, Queen Square Institute of Neurology 
  University College London \\ 
    $^8$\,The Neural Adaptive Computing Laboratory, 
  Rochester Institute of Technology
  Rochester, NY 14623  \\
  
    \texttt{tommaso.salvatori@verses.ai, ankurarjunmali@usf.edu} \\
    \texttt{chris.buckley@verses.ai, thomas.lukasiewicz@tuwien.ac.at} \\ \texttt{rao@cs.washington.edu, 
    karl.friston@verses.ai, ago@cs.rit.edu} \\
}

\begin{document}

\maketitle
\begin{abstract}
Artificial intelligence (AI) is rapidly becoming one of the key technologies of this century. The majority of results in AI thus far have been achieved using deep neural networks trained with a learning algorithm called error backpropagation, always considered biologically implausible. To this end, recent works have studied learning algorithms for deep neural networks inspired by the neurosciences. One such theory, called \emph{predictive coding} (PC), has shown promising properties that make it potentially valuable for the machine learning community: it can model information processing in different areas of the brain, can be used in control and robotics, has a solid mathematical foundation in variational inference, and performs its computations asynchronously. Inspired by such properties, works that propose novel PC-like algorithms are starting to be present in multiple sub-fields of machine learning and AI at large. Here, we survey such efforts by first providing a broad overview of the history of PC to provide common ground for the understanding of the recent developments, then by describing current efforts and results, and concluding with a large discussion of possible implications and ways forward.
\end{abstract}

\section{Introduction}
\label{sec:intro}

The machine learning community is developing and producing models that push the boundaries of the field on a weekly basis: we have witnessed considerable breakthroughs in the fields of generative artificial intelligence (AI) \cite{rombach2022high,imagen,ho2022imagen,dalle2}, game playing \cite{cicero,stratego}, and text-generation \cite{touvron2023llama,chen2021evaluating, brown20}. These results reflect over a decade of advances in the field, made possible by the joint effort of tens of thousands of researchers and engineers who have built on seminal work \cite{ciregan2012multi,Krizhevsky2012} in order to improve the performance of deep artificial neural networks trained with the error backpropagation algorithm \cite{linnainmaa1970representation,rumelhart1986learning}. %Research and application domains that have been most significantly impacted include \emph{image recognition} \cite{Krizhevsky2012,he2016deep,dosovitskiy2020vit,mixer}, \emph{speech recognition} \cite{deng2013}, \emph{game playing} \cite{silver17,moravvcik2017deepstack,vinyals2019grandmaster}, and \emph{natural language understanding} \cite{Vaswani17,devlin-etal-2019-bert,brown20}.
It may thus come as a surprise that research on alternative training methods for machine learning is more active than ever,  with many works often co-authored by the same scientists who pioneered backpropagation-based schemes \cite{bengio2015towards,scellier17,hinton2022forward,zador2022toward}.
These lines of research, however, are not \emph{contradictory} but \emph{complementary}: the increasingly impressive results of standard deep neural networks have also highlighted several significant limitations, that may be addressed via alternative training methods. There is, in fact, a common belief that, to accelerate progress in AI, we must invest in fundamental research, inspired by the findings and ideas in computational neuroscience \cite{friston2022designing,zador2022toward,da2022active}. %It seems in fact unlikely that models trained via backpropagation will be able to reach a level of intelligence, cognitive flexibility, and energy consumption that is comparable to the human brain. 
%
%Alternative methods and approaches may play an enabling role in applications where the aforementioned limitations preclude further developments and advancements. Modern deep learning models are notoriously computationally expensive \cite{dhar2020carbon,spoerer20}, untrustable \cite{marcus2018deep,marcus2019rebooting}, and biologically implausible \cite{grossberg1987competitive,crick1989recent,bengio2015towards,scellier17,whittington2019theories,ororbia20,Lillicrap20,ororbia2021towards}.
%While identifying the steps that will allow us to solve these problems is difficult, a promising direction focuses on understanding, as well as reverse engineering, learning in the human brain \cite{friston2022designing,zador2022toward,da2022active}. 
%There is, in fact, a common belief that, to accelerate progress in AI, we must invest in fundamental research in methods that are inspired by the findings and ideas in computational neuroscience \cite{friston2022designing,zador2022toward,da2022active}, as it seems unlikely that models trained via backpropagation will be able to reach a level of intelligence, cognitive flexibility, and energy consumption that is comparable to the human brain. 
%
To this end, in this review, we focus on a specific approach, inspired by a theory of learning and perception in the brain called \emph{predictive coding} (PC) \cite{rao1999predictive,friston2009predictive,millidge2022predictive}. Models of PC exhibit several valuable properties, the main one being the locality of the operations, meaning that every update of the parameters is performed using only the information of adjacent neurons or synapses in a Hebbian way \cite{hebb49}. Differently from backprop, which is a sequential algorithm, locality allows for a full parallelization of the operations, making it potentially more suitable for applications in neuromorphic computing. Locality also allows the training of artificial neural networks with any given topology \cite{ororbia20,salvatori2022learning}.  This facilitates the training of models with arbitrarily entangled structures, similar to the biological networks that constitute our brains and underwrite our level of intelligence \cite{avena18}. Furthermore, PC networks have shown to be robust and stable due to the relaxation process that spreads error signals across the whole network, and emulates implicit gradient descent \cite{song2022inferring,alonso2022theoretical,innocenti2023understanding}, allowing them to perform well in tasks that are more likely to be faced by natural agents.%, while regularizing the underlying energy landscape \cite{sengupta2018robust} and avoiding troublesome phenomena, such as vanishing and exploding gradients \cite{song2022inferring,innocenti2023understanding}. %Summarizing, PC has a solid mathematical formulation, with roots in signal compression and Bayesian inference \cite{marino2022predictive}. 

\paragraph{Notation.} In this paper, the symbol $\odot$ is employed to represent the Hadamard product, i.e., element-wise multiplication. $\oslash$ is used to indicate element-wise division. On the other hand, the symbol $\cdot$ denotes the (dot product) multiplication of matrices or vectors. The transpose of a vector $\mathbf{v}$ is represented as $(\mathbf{v})^{\mathsf{T}}$. It is important to note that matrices and vectors are distinguished by using boldface type, as in matrix $\mathbf{M}$ and vector $\mathbf{v}$, while scalars are depicted in italicized font, as in scalar $s$.  In the context of neural networks, an input observation, or a sensory input pattern, is represented as $\mathbf{o} \in \mathbb{R}^{J_0}$, where $J_{\ell}$ represents the dimension of layer $\ell$ of the network, $\ell = 0$ is the input layer and $\ell = L$ is the output layer; a label is represented by $\mathbf{y} \in \mathbb{R}^{J_{L}}$, and a latent vector (or code) by $\mathbf{x}^\ell \in \mathbb{R}^{J_\ell}$.

\section{Generative Models and Predictive Coding}
\label{sec:gen_models}

\begin{figure}[t]
    \centering
    \includegraphics[width=1.0\columnwidth]{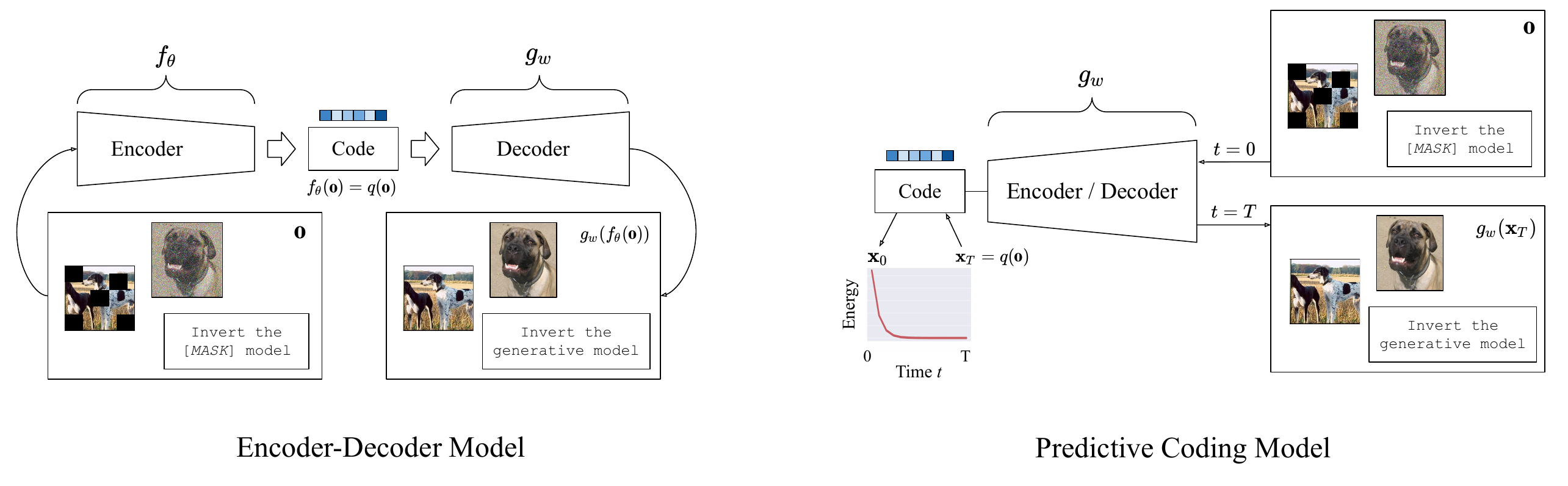}
    \caption{Generative models effectively compress information about a specific data point $\mathbf{o}$, with missing information, into a low-dimensional code vector (or latent \emph{embedding}), and use it to generate a (e.g., semantically) similar, complete data point. Left: the standard encoder-decoder model used in machine learning \cite{kingma2013auto}; Right: an equivalent PC model, which iteratively computes and refines the code through a free-energy minimization process. At convergence, the code is then used to generate a data point via the same model used to perform the compression. Typically, PC associates the encoder with ascending (prediction error) messages and the decoder with 
    descending (prediction) messages. Note that the implicit conflation of the encoding and decoding means that there is only one set of parameters in PC. These are the parameters of the generative model that, crucially, can be optimized locally, given the requisite predictions and prediction errors.}
    \label{fig:generative}
% Link to the figure:
% https://docs.google.com/drawings/d/1QQYoLU8MkpeAzNVnl7WzBeTk65erxXwDfGy7VGUhCec/edit?usp=sharing
\end{figure}

Generative modeling provides a principled statistical framework for understanding complex data by decomposing observations into latent causes and observable effects. At its foundation, a generative model specifies a joint probability distribution over observed data $\mathbf{o}$ and latent variables $\mathbf{x}$, formally expressed as $p(\mathbf{o}, \mathbf{x}) = p(\mathbf{o} \mid \mathbf{x})p(\mathbf{x})$. This factorization separates the generative process—how latent causes produce observations through the likelihood $p(\mathbf{o} \mid \mathbf{x})$—from prior beliefs about the structure of those causes, encoded in $p(\mathbf{x})$. To build intuition, consider modeling handwritten digits: the latent variables $\mathbf{x}$ might capture high-level concepts like digit identity, writing style, or stroke thickness, while the likelihood $p(\mathbf{o} \mid \mathbf{x})$ describes how these abstract concepts combine to produce the pixel intensities we observe in an image. The prior $p(\mathbf{x})$ encodes beliefs about which combinations of these concepts are more or less likely to occur naturally.

The power of generative modeling lies in its ability to address two fundamental computational problems. The first is \textit{inference}: given an observation $\mathbf{o}$, what are the most plausible latent causes that generated it? This requires computing the posterior distribution $p(\mathbf{x} \mid \mathbf{o})$ using Bayes' rule. The second is \textit{learning}: given a dataset of observations, how should we adjust the model parameters to better capture the underlying data distribution? This typically involves maximizing the model evidence $\log p(\mathbf{o})$ with respect to the model parameters, integrating over all possible latent variable configurations. In practice, both inference and learning present significant computational challenges: computing the posterior $p(\mathbf{x} \mid \mathbf{o})$ requires evaluating the often intractable marginal likelihood $p(\mathbf{o})$, while learning requires optimizing the model evidence $\log p(\mathbf{o})$ over parameter space. These computational bottlenecks have motivated the development of various approximation schemes, each making different trade-offs between accuracy and computational efficiency.

\paragraph{Variational Free-Energy.} The ability to memorize a training set, learn its patterns, and generalize to an unseen test set is the primary goal of any machine learning algorithm. The variational free energy provides a tractable objective that balances two critical factors: one that encourages fitting the training dataset (training \emph{accuracy}), and one that penalizes the \emph{complexity} of the inferred latent variables, facilitating better performance on unseen data. This enforces regularity, as it facilitates convergence towards the minimally complex model that provides an accurate fit of a specific dataset, aligned with Occam's razor formulations \cite{domingos1999occam}. Given an observation $\mathbf{o}$ and an approximate posterior distribution $q_{\theta}(\mathbf{x})$ parameterized by $\theta$, the evidence lower bound (ELBO) or negative variational free energy is defined as:

\begin{equation*}
    \mathcal{L}(\theta, \mathbf{o}) = \underbrace{\mathbb{E}_{q_{\theta}(\mathbf{x})}[\log p(\mathbf{o} \mid \mathbf{x})]}_{\text{Accuracy}} - \underbrace{D_{KL}[q_{\theta}(\mathbf{x}) \| p(\mathbf{x})]}_{\text{Complexity}}
\end{equation*}

This objective provides a lower bound on the log marginal likelihood: $\log p(\mathbf{o}) \geq \mathcal{L}(\theta, \mathbf{o})$. The accuracy term encourages the model to assign high likelihood to the observed data under the inferred latent variables, while the complexity term penalizes deviations of the approximate posterior from the prior, preventing overfitting and ensuring generalization. A typical schema to maximize this bound is the expectation-maximization (EM) algorithm, where inference first finds the variational parameters $\theta$ that maximize $\mathcal{L}(\theta, \mathbf{o})$, and then the generative model parameters are updated to follow the same goal.

\paragraph{Approximate Posterior.} Today's most prominent approach, the variational autoencoder (VAE) \cite{kingma2013auto}, uses amortized inference to compute the approximated posterior. This means that $q_{\theta}(\mathbf{x})$ is the output of the forward pass of a specifically trained neural network encoder, as sketched in Fig.~\ref{fig:generative} (left). However, an alternative class of methods performs inference iteratively, without relying on a separate encoder network. Predictive coding (PC) represents one such approach, using iterative optimization to approximate the posterior while maintaining the same generative model throughout both inference and learning, as sketched in Fig.~\ref{fig:generative} (right). This computational framework, while having roots in signal processing and neuroscience, provides a principled approach to generative modeling that we now describe in more detail.

\subsection{What Defines Predictive Coding?}
\label{sec:pc_definition}

In the PC framework, given an observation $\mathbf{o}$, the goal is to approximate the posterior distribution $p(\mathbf{x} \mid \mathbf{o})$ through iterative optimization rather than amortized inference. Historically, this process was developed as a compression mechanism, where compression is defined as the process of assigning a lower-dimensional latent code to a particular observation. The first algorithmic formulation of PC dates back to the 1950s as a compression algorithm for time series data \cite{cutler54, o1971entropy, elias1955predictive}, and it is this \emph{coding} that gave the framework its name. The first instances of PC in neuroscience also treated it as a pure inference algorithm, with no learning involved, as the goal was to model how spatial information was encoded in the retina \cite{srinivasan1982}.

\begin{figure}[t]
    \centering
    \includegraphics[width=1.0\columnwidth]{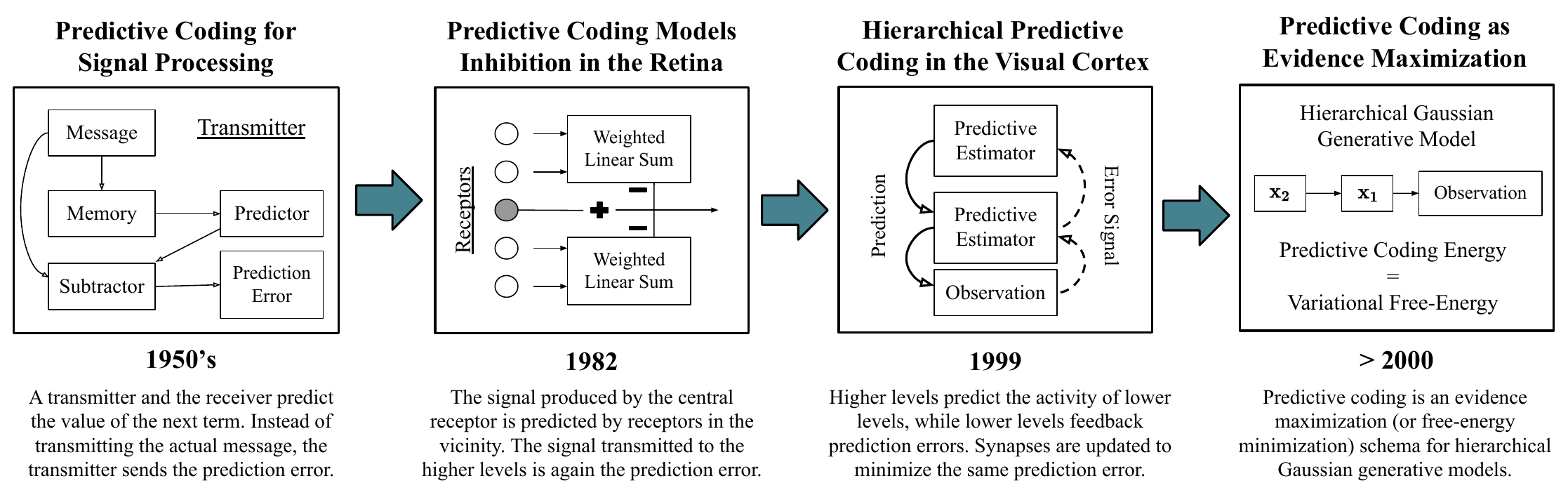}
    \caption{A timeline of how the perception of what PC is has changed through the years. Initially, it was developed as a signal compression mechanism \cite{elias1955predictive,elias1955predictiveII}; then, it was used to model inhibition in the retina \cite{srinivasan1982}. It then became a more general model of both learning and perception in the visual cortex \cite{rao1999predictive}. Nowadays, it can be abstractly  defined as an evidence-maximization scheme for hierarchical Gaussian generative models \cite{friston2005theory,friston2009predictive}. For a detailed discussion on different PC algorithms, we also refer to another survey \cite{spratling2017review} }
    \label{fig:history}
% Link to the figure:
% https://docs.google.com/drawings/d/1J8buupMcVZgGNlVv00yOKGuOBc5xEuBOKGSl5ah449s/edit?usp=sharing
\end{figure}

A major paradigm shift occurred several years later with the work of Rao and Ballard, which employed PC as a \emph{learning} algorithm in addition to inference, modeling hierarchical processing in the visual cortex \cite{rao1999predictive}. In this formulation, synaptic weights are updated after inference to minimize the prediction error of the model. The connection with variational inference was established later, connecting PC to hierarchical Gaussian generative models \cite{friston2003learning,friston2008hierarchical}. As a result of these independent but deeply related lines of work, a general theory has unified these views by showing that iterative updating schemes can be described as a process of minimizing a \emph{variational free energy} \cite{friston2010}, also known in machine learning as the evidence lower bound (ELBO) \cite{winn2005variational}. A sketch of the timeline of main breakthroughs in PC history is given in Fig.~\ref{fig:history}.

The PC literature has historically focused on a specific class of models that make two key simplifying assumptions: first, it typically assumes that all probability distributions are multivariate Gaussian; second, it often employs a hierarchical structure where latent variables follow a first-order Markov assumption—each layer depends only on the layer immediately above it. These assumptions, while restrictive, provide significant computational advantages: Gaussian distributions are conjugate to themselves under linear transformations, enabling closed-form updates, and the Markov structure allows for local computations between adjacent layers.% It is worth noting that recent work has begun to relax these assumptions, exploring non-Gaussian distributions and more complex dependency structures, but the classical PC framework provides a principled starting point \cite{pinchetti2022, salvatori2022learning}.

Let $\mathbf{o} = g(\mathbf{x},\mathbf{w})$ be a generative model, where a vector of latent variables $\mathbf{x}$ and a set of parameters $\mathbf{w}$ are used to generate an observation $\mathbf{o}$. In our case, we consider generative models with $L$ layers, whose joint probability follows the hierarchical dependencies:
\begin{align}
    p(\mathbf{x}_L, \dots, \mathbf{x}_0) = p(\mathbf{x}_L) \prod_{\ell=0}^{L-1} p(\mathbf{x}_\ell \mid \mathbf{x}_{\ell+1}).
\end{align}
We further consider each conditional probability distribution $p(\mathbf{x}_\ell \mid \mathbf{x}_{\ell+1})$ to be a multivariate Gaussian distribution, whose mean is given by a transformation $g_\ell$ of the latent variables of the level above. In the majority of the literature, the map $g_\ell$ is a composition of an activation function (such as ReLU), and a transformation matrix $\mathbf{w}_\ell$ (with bias $\mathbf{b}_\ell$ possibly folded into this matrix), which results in a linear map. As a result, we then, more formally, arrive at the following:
\begin{align}
    p(\mathbf{x}_L) = \mathcal{N}(\mathbf{x}_L; \boldsymbol{\mu}_L, \boldsymbol{\Sigma}_L), \quad p(\mathbf{x}_\ell \mid \mathbf{x}_{\ell+1}) = \mathcal{N}(\mathbf{x}_\ell; g_{\ell+1}(\mathbf{x}_{\ell+1}), \boldsymbol{\Sigma}_\ell).
\end{align}
\paragraph{Inversion.} Given the above, we now have the structure of the main object of study that centrally characterizes the PC literature, that is, a hierarchical generative model, which lives in a continuous state space and whose probability distributions are Gaussian in nature. However, what also defines PC is the process that is used to invert the generative model, which means estimating $p(\mathbf{x}_1, \dots, \mathbf{x}_L \mid \mathbf{o})$ via an approximate posterior $q(\mathbf{x}_1, \dots, \mathbf{x}_L)$. To do this, we need to appeal to two different approximation results: first, we leverage a mean-field approximation, which allows us to assume that the variational posterior factorizes into conditionally independent posteriors $q(\mathbf{x}_\ell)$; then, through the Laplace approximation, we assume that approximate posterior distributions are Gaussian in form \cite{friston2008variational}. At this point, the model can be inverted by minimizing the resulting variational free energy via gradient descent or fixed-point iterations.

As the resulting variational free energy has a quadratic form \cite{millidge2020predictive, friston2005theory}, its gradients correspond to the linear and weighted prediction errors defined in the original computational model by Rao and Ballard \cite{rao1999predictive}, which was notably developed before PC was cast as variational learning and inference \cite{friston2005theory}. We have now introduced all of the core concepts needed to provide a  definition of PC. This definition could be used as an umbrella for all of the variations on PC in the literature.
\begin{definition}[Informal]
    Let us assume that we have a hierarchical generative model $g(\mathbf{x},\mathbf{o})$, inverted using an algorithm $\mathcal A$. Then, $\mathcal A$ is a predictive coding algorithm if and only if:
    \begin{enumerate}[noitemsep,nolistsep]
        \item it maximizes the model evidence  $\log p(\mathbf{o})$ by minimizing a variational free energy, 
        \item the posterior distributions of the nodes of the hierarchical structure are factorized via a mean-field approximation, and 
        \item each posterior distribution is approximated under the Laplace approximation (i.e., random effects are Gaussian).
    \end{enumerate}
\end{definition}
Note that the above definition does not say anything explicitly about \emph{prediction error} or properties such as \emph{locality}, which, as mentioned earlier, are commonly used to describe PC. This is because they are not foundational to PC but rather consequences of the commitment to the aforementioned generative model: the mean field approximation enforces independence, and hence, results in locality in the update rules; the Laplace approximation simplifies variational free energy to a quadratic function, meaning that its gradients are linear prediction errors.

It is crucial to note that the above definition of PC does not mean, however, that the algorithms surveyed here are optimal and cannot be improved: the above definition is quite general and does not impose any constraint on the exact computation of the posteriors as well as the optimization  technique(s) used to minimize the variational free energy. It also does not mean that such conditions cannot be relaxed: recent works have also explored alternatives to Gaussian and Markov assumptions. Common alternatives use amortized computations using a similar technique to  VAEs \cite{vafaii2024poisson,vafaii2025brain, zahid2023sample, tscshantz2023hybrid}, rely on different probability distributions \cite{vafaii2025brain,pinchetti2022} or use models with different architectures \cite{salvatori2022learning,salvatori2023predictive}. It is also worth noting a recent trend that relies on sampling algorithms to improve the generative capacities of PC models \cite{sennesh2024divide,oliviers2024learning, zahid2023sample}. However, we believe the classical framework remains essential for understanding the foundational principles that give PC its characteristic properties. 

%Moreover, the mathematical tractability provided by these assumptions continues to make this formulation valuable for theoretical analysis and serves as a stepping stone for developing more sophisticated variants. The definition thus provides a unifying conceptual foundation that helps organize the diverse landscape of modern PC approaches around their core computational commitments.

\section{Implementations of Predictive Coding}
\label{sec:pc_frameworks}

\begin{figure}[t]
    \centering
    \includegraphics[width=1.0\columnwidth]{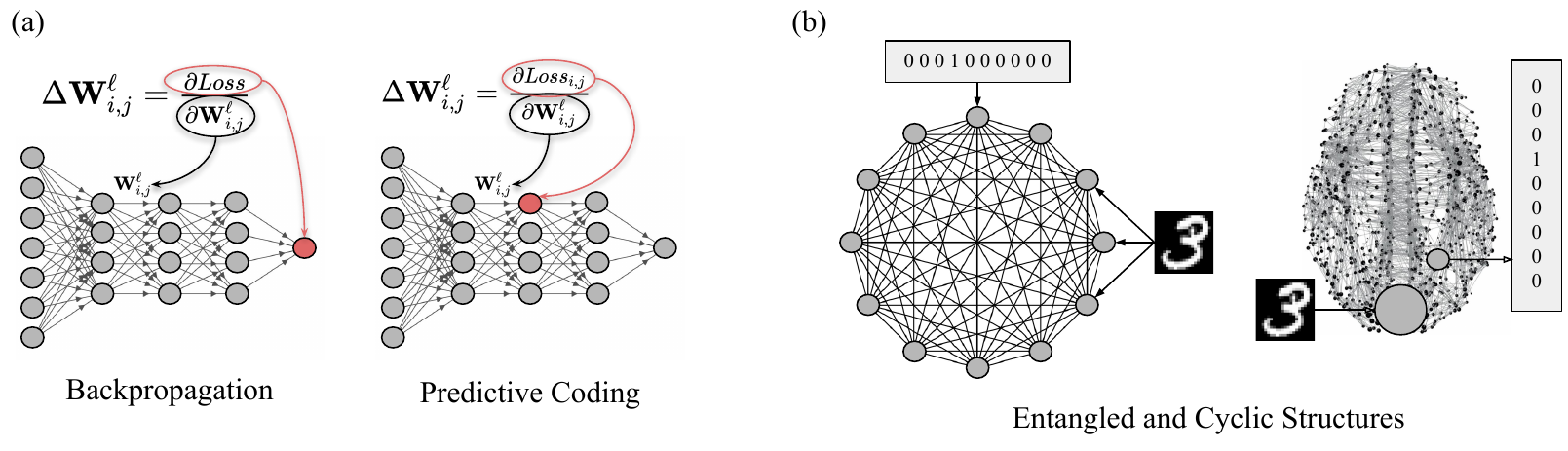}
    \caption{(a) The difference between PC and standard models in terms of locality: backprop updates its synaptic weights $\w^{\ell}_{i,j}$ to minimize the output error, even if it is not directly connected to it. PC models, on the other hand, perform their updates to correct the error of their postsynaptic neuron. (b) PC can be used to train models with cycles. An example is the fully connected model on the left, where every pair of neurons is connected via two different synapses, one in each direction. Being able to train models like this facilitates the inversion of models with arbitrarily expressive architectures (such as realistic brain structures) by simply masking specific connections of a fully connected model via an adjacency matrix. For an example of interesting models that operate on realistic brain structures, see right panel, taken from \cite{avena18}.
    }
    \label{fig:pc-bp-basic}
\end{figure}

%In the following section, we will concretely explicate the framework of fundamental PC (PC) as well as define the dynamics of two of its prominent and popular formulations found in the literature:  %because of ABC

%Given an understanding and framing of backprop, 
%We now turn attention to this survey's core credit assignment framework of interest: PC (PC), also known as predictive processing \cite{clark2015surfing}.
%Let us assume we have a PC model, in shape identical to a standard neural network, with $L$ layers of dimension $J_{\ell}$, and tensors of weight parameters $\mathbf{W}^{\ell+1}$.
Informally, PC claims that there exist two families of neurons that define an internal (i.e., generative) model of the world: the first generates predictions that are passed to lower layers, and  the second encodes prediction errors that are passed to higher layers \cite{rao1999predictive,whittington2017approximation}. Recent work has applied this basic separation to prediction and prediction errors within the compartments of a single neuron, where errors are propagated back via dendritic connections \cite{sacramento2018dendritic, whittington2019theories}. In this work, we use the standard formulation of PC papers in machine learning, and describe an artificial neuron as a computational unit with three quantities: its value (\emph{value node}), its \emph{prediction}, and its error (\emph{error node}), defined as the difference of the first two. 
The \emph{value node} $\x^{\ell}_{i,t}$ (where the indices $i$ and $\ell$ refer to the $i$-th neuron of the $\ell$-th layer at time $t$) encodes the most likely value of some latent state. The second computational unit is the \emph{prediction} $\mathbf{u}^{\ell}_{i,t}$, which is a function of the value nodes of a higher level in the hierarchy. Let $J_{\ell}$ denote the dimension (number of neurons) of layer $\ell$, $\mathbf{W}^{\ell+1}$ be a matrix containing the predictive synaptic parameters for layer $\ell+1$, and $\phi$ be a nonlinear activation function. Then, the prediction is computed as follows:
\begin{equation}
\mathbf{u}^{\ell}_{i,t} = \sum_{j=1}^{J_{\ell}} \mathbf{W}^{\ell+1}_{i,j} \phi( \mathbf{x}^{\ell+1}_{j,t} ),
\label{eq:pcn-forward}
\end{equation}
where $\phi$ is a nonlinear function (akin to the activation function in deep neural networks). The third computational unit is the prediction error $\mathbf{e}^{\ell}_{i,t}$, given by the difference between its value node and its prediction node, i.e., $\e^{\ell}_{i,t} = \mathbf{x}^{\ell}_{i,t} - \mathbf{u}^{\ell}_{i,t}$. This local definition of error, which exists in every network neuron, foregrounds a key difference between PC and models trained with backprop (e.g., the multilayer perceptron), as it enables learning through only local computations. A graphical example depicting the differences in locality between PC and backprop is given in Fig.~\ref{fig:pc-bp-basic}. Taken together, the above three quantities, as well as the set of synaptic weight matrices $(\w^0, \dots, \w^L)$, define a generative model where both inference and learning are performed as a means to minimize a single (global) energy function, formally defined as the sum of the squared prediction errors of every neuron:
\begin{equation}
    \mathcal{E}_{t} = \frac{1}{2} \sum_{i,\ell} ( \e^{\ell}_{i,t} )^2.
    \label{eq:energy}
\end{equation}
This energy function is exactly the variational free energy defined in the previous section, and a proper derivation can be found in \cite{millidge2021predictive}. Let us assume that our generative model is presented with an observation $\mathbf{o} \in \mathbb R^{J_0}$. Then, the following process describes the credit assignment and consequent update of the synaptic parameters of the model: first, the neurons of the lowest layer are set equal to the sensory observation, i.e., $\x^0 = \mathbf{o}$. Next, the unconstrained neural activities are updated until the convergence step $T$ (or, for a fixed number $T$ of iterations) to minimize the energy of Eq.~\ref{eq:energy} via gradient descent. In particular, the equation for the update dynamics of the value nodes is the following:
\begin{equation}
\!\Delta{\x}^{\ell}_{t} = - \gamma \cdot \frac{\partial \mathcal{E}_{t}}{\partial \x^{\ell}_{t}} =
\begin{cases}
\gamma \cdot (\phi' ( \x^{\ell}_{t} ) \odot (\w^{\ell})^{\mathsf{T}} \cdot \e^{\ell-1}_{t} ) & \!\!\mbox{if } \ell = L \text{ (and unclamped)}\\
\gamma\cdot ( -\e^{\ell}_{t} + \phi' ( \x^{\ell}_{t} ) \odot (\w^{\ell})^{\mathsf{T}} \cdot \e^{\ell-1}_{t} ) & \!\!\mbox{if } 0 \,{<}\ \ell \,{<}\ L\\
\gamma \cdot (-\e^{\ell}_{t} ) & \!\!\mbox{if } \ell=0\text{ (and unclamped),}
\end{cases}
\label{eq:pcn-inference}
\end{equation}
%
% Alex: I fixed this b/c this equation had a bug, probably b/c it is showing PC in what I call "bottom-to-top" format and then the figure doesn't align with it (as opposed to "top-to-bottom" format like in later equations), b/c we clamp x^0 = o, we get wrong dynamics for layer x^1 and x^0: I have fixed the issue above
% \begin{equation}
% \!\Delta{\x}^{\ell}_{t} = - \gamma \cdot \frac{\partial \mathcal{E}_{t}}{\partial \x^{\ell}_{t}} =
% \begin{cases}
% \gamma\cdot ( -\e^{\ell}_{t} + \phi' ( \x^{\ell}_{t} ) \odot (\w^{\ell})^{\mathsf{T}} \cdot \e^{\ell+1}_{t} ) & \!\!\mbox{if } 0 \,{<}\ \ell \,{<}\ L\\
% \gamma \cdot (\phi' ( \x^{\ell}_{t} ) \odot (\w^{\ell})^{\mathsf{T}} \cdot \e^{\ell+1}_{t} ) & \!\!\mbox{if } \ell=0\text{,}
% \end{cases}
% \label{eq:pcn-inference}
% \end{equation}
% %
where $\gamma$ is the learning rate of the neural activities (usually read as a precision or inverse variance). This optimization process drives the underlying credit assignment process of PC and computes the best configuration of value nodes needed to perform a synaptic weight update. When this process has converged, the value nodes are then frozen and a single weight update is performed (via gradient descent) to further minimize the same energy function:
\begin{equation}
\Delta \w^{\ell+1} = -\alpha \big( {\partial \mathcal{E}_{T}}/{\partial \w^{\ell+1}} \big)
=  \alpha \big( \e^{\ell}_{T} \cdot \phi( \x^{\ell+1}_{T})^{\mathsf{T}} \big),
\label{eq:pcn-update-param}
\end{equation}
where $\alpha$ is the learning rate for the synaptic update. The alternation of these two phases, i.e., the value node update and weight update steps, defines the learning algorithm used to train PC networks   \cite{rao1999predictive,whittington2017approximation}. Importantly, although every computation is local, both update rules minimize the same energy function, which is globally defined over the entire network, as described in \ref{algo:PC}. The difference in locality between a weight update of backprop and the one for PC is given in Fig.~\ref{fig:pc-bp-basic}(a).
\begin{algorithm}[t]
    \footnotesize
    \caption{Updating the value/state dynamics and synaptic weights in a PC model, given a data point $\mathbf{o}$.}\label{algo:PC}
    \begin{algorithmic}[1]
    \REQUIRE $\x^L$ is fixed to $\mathbf{o}$ for every time step $t$.
    \FOR{$t=0$ to $T$ (included)}
        \FOR{each neuron $i$ in each level $\ell$}
            \STATE Update $\x^{\ell}_{i,t}$ to minimize $\mathcal{E}_{t}$ via Eq.~\eqref{eq:pcn-inference}\
        \ENDFOR
        \IF{$t= T$}
            \STATE Update $\w^{\ell}$ to minimize $\mathcal{E}_{t}$ 
            via Eq.~\eqref{eq:pcn-update-param} \ \  
            {
          %  \RETURN
            }
        \ENDIF
    \ENDFOR
    \end{algorithmic}
\end{algorithm}

\begin{figure}[t]
    \centering
    \includegraphics[width=0.675\columnwidth]{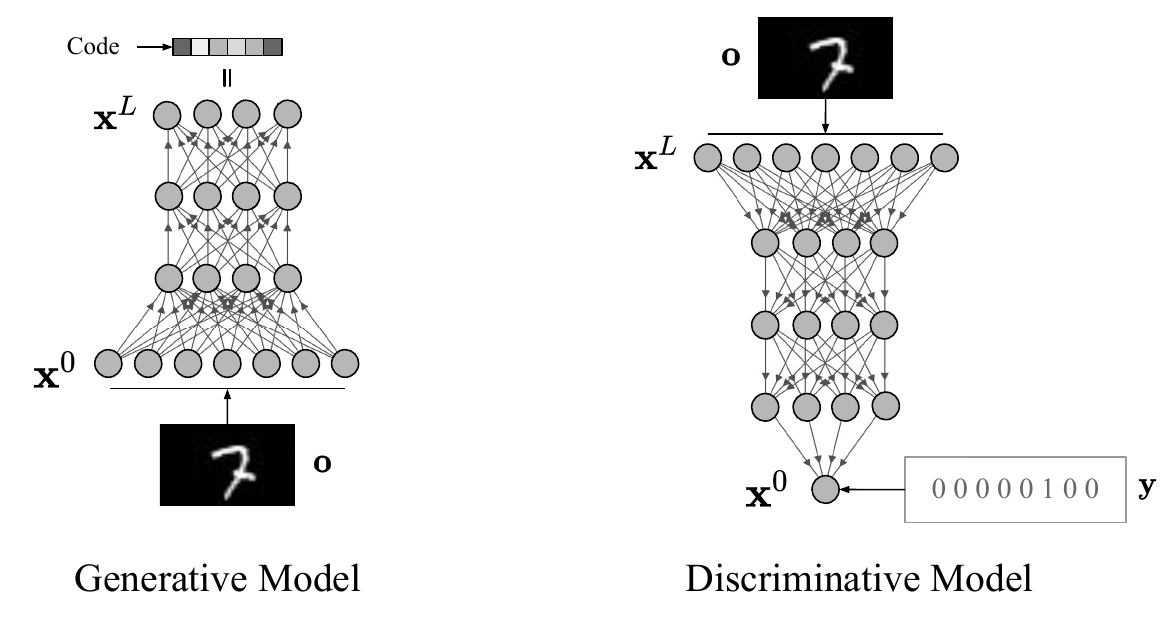}
    \caption{A graphical depiction of PC  in an unsupervised generative form (left) and in a supervised discriminative form (right). Note that the generative form of PC entails iteratively inferring a latent code (or embedding) $\mathbf{x}^L$ for sensory input $\mathbf{x}^0 = \mathbf{o}$, while the discriminative form of PC requires iteratively learning a predictive mapping between sensory input $\mathbf{x}^L = \mathbf{o}$ to target $\mathbf{x}^0 = \mathbf{y}$.
    }
    % https://docs.google.com/drawings/d/140AmHLD6TbSELz_Mdi_uS5niDgOC7LxZBHBJWoBjcUQ/edit
    \label{fig:pc-model-types}
\end{figure}

\paragraph{Supervised Learning.}
In the above, we have described how to use PC to perform unsupervised learning tasks. However, this algorithm  or variations of it have been shown to obtain a performance comparable to deep networks learned with backprop on supervised learning tasks \cite{whittington2017approximation,salvatori2022incremental,han2018deep}. To extend the above formulation to the domain of labeled data, we have to re-imagine the PC network as \emph{generating the label} $\mathbf{y}$ and that the data point $\mathbf{o}$ serves as a prior on the value of the neurons in the first layer of the network. Let us assume that we have a labeled data point $(\mathbf{o},\mathbf{y})$, where, again, $\mathbf{y}$ is the label
\footnote{In supervised learning, the total energy of the model can also be decomposed as
    $\mathcal{E} = \tilde{\mathcal{E}_{t}} + \mathcal{L}$, 
 with  $\tilde{\mathcal{E}_{t}}$ be the total energy of all the internal neurons only, and $\mathcal{L}$ be the squared error defined on the label, as in standard regression tasks. This allows to define interesting similarities with backprop \cite{millidge2022theoretical}.}. 
Supervised learning is then realized by fixing the value nodes of the first and last layer to the entries of the data point and its label, respectively, i.e.,  $\x_{0,t} = \mathbf{o}$ and $\x_{L,t} = \mathbf{y}$ for every time step $t$. A graphical depiction of the difference between supervised and unsupervised learning using PC is given in Fig.~\ref{fig:pc-model-types}.

\paragraph{Generalization to Arbitrary Topologies.} 
It is often remarked that the networks in the brain are not strictly hierarchical \cite{ororbia20,byiringiro2022robust} but are extremely entangled and full of cyclic connections. Recent work has shown how PC can be utilized to model networks with any kind of structure, making it ideal to digitally perform learning tasks that require brain-like architectures such as parallel cortical columns or sparsely connected brain regions \cite{salvatori2022learning}. To do this, the more general concept of a PC graph is required. Let $G=(V,E)$ be a fully connected directed graph, where $V$ is a set of $n$ vertices $\{1,2,\dots,n\}$ and $E\subseteq V\times V$ is a set of directed edges between them, where every edge $({i,j})\in E$ has a weight parameter encoded in an $n \times n$ weight matrix $\w$. As a result, the predictions come from all of the neurons of the network, and the free energy of the model is again the summation over all of the (squared) prediction errors:
\begin{equation}
     \u_{i,t} = \sum_{j=0}^n \w_{j,i} \phi(\x_{j,t}) \ \ \  \ \ \ \ \text{and} \ \ \ \ \ \ \ \mathcal{E}_{t} = \frac{1}{2} \sum_{i=0}^n ( \e_{i,t} ) ^2.
    \label{eq:arbitrary_energy}
\end{equation}
While defining this generalization on fully connected graphs makes the notation simpler, interesting use cases can be found in graphs that are not fully connected. To this end, it suffices to note that every graph is a subset of a fully connected one and, hence, we can consider sparse weight matrices $\w$, where only the entries of the edges that we need to define our structure are non-zero. This can be done by multiplying $\w$ by an adjacency matrix. In the original work along these lines, the authors trained a network of multiple, sparsely connected brain regions and tested the resulting model on image denoising and image completion tasks. The exact structure of this network is that of the \emph{Assembly Calculus}, a Hebbian learning framework/method specifically designed to model brain regions \cite{Papadimitriou20}. A graphical sketch of how to derive such a model is given in Fig.~\ref{fig:pc-bp-basic}(b). 

\paragraph{Associative Memories.} A body of literature has investigated the efficacy of PC networks in memory-related tasks. Specifically, this literature studies their capacity to store a dataset and retrieve individual data points when a related cue is presented. In practice, this cue is often an incomplete or corrupted version of a stored memory, and the task is successfully completed if and only if the correct data point is retrieved. PC schemes have demonstrated their ability to effectively store and retrieve complex memories, such as images from the COIL and ImageNet datasets \cite{rao1999visionresearch,salvatori2021associative,tang2023recurrent}. However, while their retrieval capabilities are robust, their capacity is still limited, and hence not comparable to that of continuous or universal Hopfield networks \cite{millidge2022universal, ramsauer21}. This capacity can be improved by implementing a fast and powerful memory writing operation that allows these models to store individual memories without overwriting existing ones. Models that incorporate memory writing also tend to facilitate the implementation of a \emph{forget} operation that erases individual memories with little to no impact on overall model performance \cite{yoo2022bayespcn}. It is also possible to store memories in the latent variable $\x_L$, by learning a Gaussian mixture prior, whose centers correspond to each of the individual memories \cite{annabi2022relationship}. This has also been implemented in the temporal domain, where the goal is to retrieve missing future frames of a video, given initial ones \cite{tang2023recurrent} (as a sort of priming stimulus). Note, however, that this class of models does not have the exponential capacity of modern Hopfield networks \cite{ramsauer21,Krotov21}.

\subsection{Differences and Similarities with Backpropagation}

In the case of supervised learning, PC can be seen simply as a training algorithm for deep neural networks that can be used as an alternative to backprop. However, the two algorithms differ in terms of convergence, complexity, and performance, in ways that we now discuss.

\paragraph{Similarities.} Recent work has shown that PC can approximate the weight update of backprop, under specific conditions, on both MLPs (multilayer perceptrons) and within the more general framework of computational graphs \cite{whittington2017approximation, millidge2020predictive, millidge2022backpropagation, rosenbaum2022relationship}. These conditions are restrictive in practice, as these results only hold if either the total prediction error on the network is infinitesimally small or the predictions are kept constant throughout the whole duration of the inference process, i.e., $\u^{\ell}_{i,t} = \u^{\ell}_{i,0}$ for every time step $t$. In practice, however, empirical studies have shown that for the approximation to hold, it suffices to have a small output error \cite{millidge2020predictive}. A similar line of work has also shown that simply adding a temporal scheduling to the updates of the weights and the neural activities leads to a weight update that is equivalent to that of backprop \cite{Song2020,salvatori2022reverse}. This temporal scheduling, that sequentially  updates the weights of different layers at different, pre-defined, time steps, is however implausible, as it requires external control that triggers the updates at different time steps \cite{zahid2023predictive}. 

%Similar works have also investigated similarities between PC and other neuroscience-inspired algorithms, such as contrastive Hebbian learning \cite{movellan1991contrastive}, target propagation \cite{bengio2014auto,lee2015difference,millidge2022backpropagation}, and equilibrium propagation \cite{scellier17,millidge2022theoretical}.

\paragraph{Differences.}
PC has been shown to perform better than standard models (such as backprop-trained deep neural networks) on problems that are faced by biological organisms, such as continual learning, online learning, and learning from a small amount of data \cite{ororbia2019lifelong,ororbia2020continual,song2022inferring}. This is due to the inference phase, which allows the error to be distributed in the network in a way that avoids a phenomenon called \emph{weight interference}. A second difference is about stability and convergence: in backprop, and in models trained with gradient descent on a non-local loss function, each parameter update is computed independently based on the current network state, without accounting for how other parameters will simultaneously change. This has been shown to produce updates that destabilize training \cite{barrett2020implicit}. This is not true for PC networks, as it has been shown that PC models trained for supervised learning naturally implement \emph{implicit} gradient descent \cite{alonso2022theoretical}, a more stable optimization approach where each parameter update takes into account the simultaneous changes that will occur to all other parameters in the network. Specifically, a parameter update of implicit gradient descent is defined as follows:
\begin{align}
    \w_{t+1} = \w_t - \gamma \nabla \mathcal L(\w_{t+1}).
\end{align}
This formulation is called \emph{implicit},  since $\w_{t+1}$ appears on both sides of the equation, desirably reducing the sensitivity to the learning rate \cite{toulis2014statistical}. As a consequence, PC models tend to be more robust and better calibrated than standard ones, as has been shown in multiple classification tasks on both convolutional and graph networks \cite{byiringiro2022robust,choksi2021predify,salvatori2022incremental}. Recent works have also studied and analyzed the convergence of PC models, showing that the iterative inference phase allows them to better escape bad local minima and saddle points, as well as needing a smaller number of weight updates with respect to backpropagation \cite{innocenti2023understanding,mali2024tight}.

\paragraph{Limitations.} Differently from backprop-based models, PCNs face a scalability bottleneck that limits their effectiveness on large-scale deep learning tasks, preventing any kind of industrial adoption. For example, in classification tasks, these models achieve competitive performance  when tested on standard benchmarks with shallow architectures, such as convolutional models with 5-7 layers. However, we observe that performance degrades significantly as network depth increases \cite{pinchetti2024benchmarking}. This has been shown both in PC models trained with gradient descent, and in models trained via closed-form updates \cite{koudhal25}. In both cases, the core issue stems from energy propagation dynamics: deeper networks experience exponentially larger energy magnitudes in layers closer to the output, creating unstable training conditions where parameter updates concentrate in a small subset of layers rather than utilizing the full network depth.
This energy imbalance causes latent states to diverge excessively from forward pass activations, leading to suboptimal weight updates and poor credit assignment throughout the network. While strategies such as energy regularization, modified weight update schemes, and initialization strategies are improving the results \cite{innocenti2025mu,qi2025training,goemaere2025error}, there is still a significant amount of research to be performed to be able to make PC match the performance of BP on larger architectures, such as language models \cite{devlin-etal-2019-bert}.

A second limitation concerns the computational complexity of the inference phase. Theoretically, optimal performance requires waiting until convergence before updating weights, which introduces substantial computational overhead. In practice, however, a limited number of inference iterations yields competitive results, typically requiring iterations at least equal to the network depth, with empirical evidence suggesting that optimal performance occurs when $L < T < 2L$ iterations are performed \cite{pinchetti2024benchmarking}. When the iteration count approximates the layer count, the time complexity becomes comparable to backprop, though space complexity remains higher due to the need to maintain all neurons and layers simultaneously in memory. This computational profile creates significant engineering challenges, particularly the absence of deep learning libraries capable of fully parallelizing the inference phase. Current implementations fall into two distinct categories: libraries that achieve full parallelization but lack seamless integration with established deep learning frameworks like PyTorch and Equinox, such as NGC-Learn \cite{ororbia20} and jPC \cite{innocenti2024jpc}; and libraries that provide modular, framework-compatible approaches but sacrifice parallelization, such as PCX \cite{pinchetti2024benchmarking}.

\subsection{Neural Generative Coding}

\begin{figure}[!t]
\centering
\includegraphics[width=6.in]{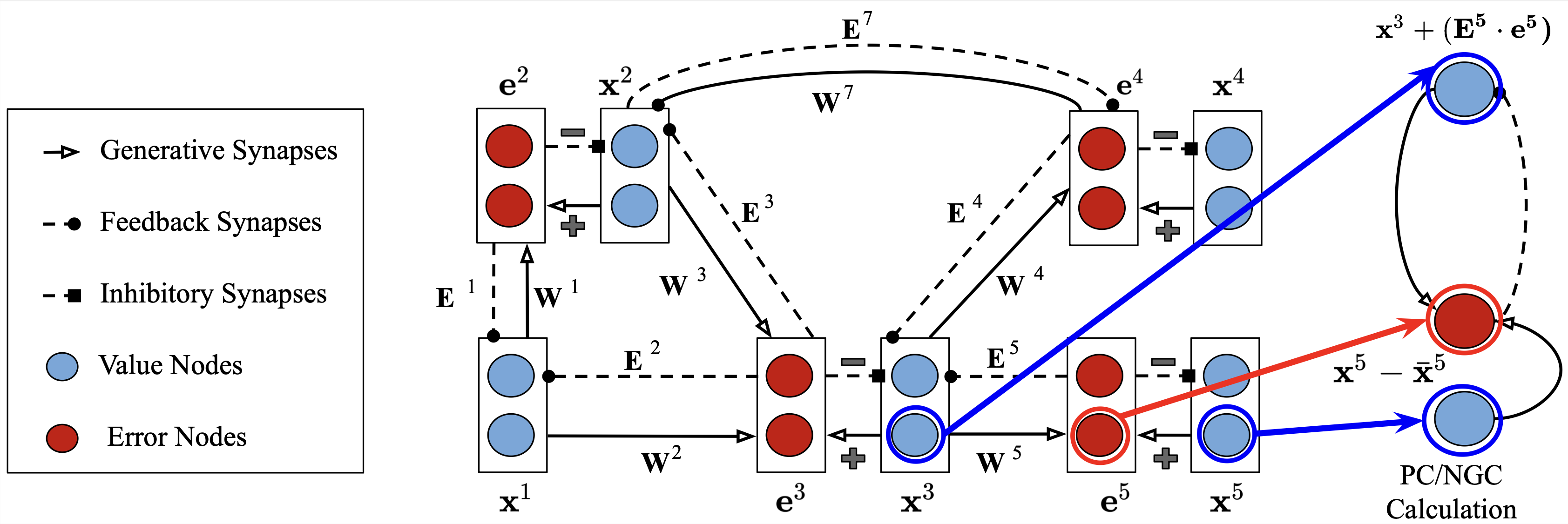}
\caption{On the left is an arbitrary neural circuit (non-linearities are omitted for simplicity) that could be constructed under one of the three PC frameworks surveyed. Rao \& Ballard PC (PC) only employs generative synapses $\mathbf{W}^{\ell}$, while NGC further employs separate feedback synapses (which may or may not correspond symmetrically to the generative pathway). Note that NGC/BC-DIM notably often employ skip connections ($\mathbf{W}^7$). %Finally, the right (which zooms in to a single value-error-value node triplet sub-circuit) depicts core differences in internal operations employed by PC/NGC and BC-DIM: PC/NGC computes error nodes via subtraction and updates value nodes using addition, while BC-DIM computes error nodes via division and updates value nodes using multiplication.
}
%\vspace{-0.5cm}
\label{fig:pc_circuit_variants}
\end{figure}

Despite being developed to model information processing in different brain areas, the above formulation is still biologically implausible, the main implausibility being symmetric weight connections. In what follows, we survey a generalization of PC, to arbitrary wiring patterns, called neural generative coding (NGC). This variation has its roots grounded in cable theory \cite{hodgkin1946electrical,rall1962theory} and neuronal compartments \cite{ermentrout2010mathematical,hodgkin1946electrical,rall1962theory}. Like PC, NGC employs predict-then-correct learning but incorporates additional neurobiological mechanisms including lateral competition, learned precision-weighting, and membrane potential leakage. The key innovation of NGC is decoupling the forward generative pathway (matrices $\mathbf{W}^\ell$) from the feedback pathway (matrices $\mathbf{E}^\ell$). This allows flexible connectivity patterns, including skip connections, without requiring that feedback structures mirror the generative model in reverse \cite{ororbia20}. Each NGC circuit processes input over $T$ time steps, where $g^\ell$ represents a precision-weighting function, $\beta$ controls the update rate, $\gamma$ represents membrane leakage, $f_D$ is a dendritic processing function, $\Delta t$ is the time step size, and $\Phi(\mathbf{x}^\ell_t)$ captures lateral interactions. Predictions are computed and neural activities are updated according to:
\begin{align}
&\mathbf{\bar{x}}^\ell_t = g^\ell \Big( \mathbf{W}^{\ell+1} \cdot \phi^{\ell+1}( \mathbf{x}^{\ell+1}_t ) \Big), \\ 
&\mathbf{x}^\ell_{t+\Delta t} = \mathbf{x}^\ell_t + \beta \Big(-\gamma \mathbf{x}^\ell_t -\mathbf{e}^\ell_t + (\mathbf{E}^\ell \cdot \mathbf{e}^{\ell-1}_t) \odot f_D( \mathbf{x}^\ell_t )  + \Phi(\mathbf{x}^\ell_t) \Big).\label{eqn:state_update}
\end{align}
Error signals are calculated as precision-weighted differences between actual and predicted activities, modulated by cross-correlational matrices $\mathbf{\Sigma}^\ell$.
After settling for $T$ steps, both generative and feedback synapses are updated via multi-factor Hebbian rules with $\gamma_e < 1$ ensuring feedback synapses evolve more slowly than generative ones:
\begin{align}
&\Delta \mathbf{W}^\ell = \mathbf{e}^\ell_t \cdot (\phi^{\ell+1}( \mathbf{x}^{\ell+1}_t) )^T, \\
&\Delta \mathbf{E}^\ell = \gamma_e ( \phi^{\ell+1}(\mathbf{x}^{\ell+1}_t) \cdot (\mathbf{e}^\ell_t)^T ).
\end{align}
For a detailed explanation of all the mechanisms of NGC, we refer to the supplementary material.

\section{Predictive Coding in Machine Learning}
\label{sec:pc_applications}

Despite the limitations regarding the scale highlighted in the previous chapter, PC has demonstrated remarkable versatility across a large number of machine learning tasks. While early implementations focused primarily on simple classification tasks, recent advances have shown that PC can effectively tackle increasingly sophisticated problems that were previously dominated by backpropagation-based approaches. These developments that we now survey, highlight both the practical potential and current limitations of predictive coding as a comprehensive learning paradigm for modern machine learning applications.

\paragraph{Supervised Learning.}
The first application of PC to supervised learning involved training a small PC network to perform image classification on the MNIST dataset, achieving test and train errors comparable to those of an MLP of the same complexity (depth and width) \cite{whittington2017approximation}. Since then, similar results have been achieved on convolutional networks trained on datasets of \emph{RGB} images, such as CIFAR10 and SVHN \cite{salvatori2022incremental}. In this work, the authors showed that updating the weight parameters alongside the neural activities (i.e., running Equations \ref{eq:pcn-inference} and \ref{eq:pcn-update-param} in parallel), instead of waiting for the inference phase to converge, improves test accuracies as well as ensures better convergence. These results also extend to structured datasets and graph neural networks, where PC is again able to match the performance of backprop on different benchmarks, with the advantage of learning models that are better calibrated and more robust to adversarial examples \cite{byiringiro2022robust}.   

\paragraph{Natural Language Processing.} Gaussian assumptions (that underwrite PC) can be limiting in scenarios where we need to model different distributions, such as categorical distributions or mixture models. One such scenario is found in transformer models \cite{Vaswani17}: the attention mechanism encodes a categorical distribution, computed via the softmax activation. To this end, it is possible to generalize the definition of the energy of a specific layer, originally defined as the squared distance between the predictions $\u^{\ell}$ and the activities $\x^{\ell}$, to the KL divergence between two probability distributions, with predictions and activities serving as sufficient statistics. In detail, the PC model now has the following variational free energy:
\begin{align}
\mathcal{E}_{t} = \sum\nolimits_{\ell=0}^{L} \mathcal{E}^{\ell}_t \ \ \ \ \text{where} \ \ \ \ \mathcal{E}^{\ell}_t = D_{KL}[\mathcal{X}^{\ell}(\u^{\ell}_t) || \mathcal{X}^{\ell}(\x_t)],
\label{eq:fkl}
\end{align}
where $\mathcal{X}^{\ell}(x)$ is a probability distribution defined for every layer with sufficient statistics $x$. If every distribution is a Gaussian, this is equivalent to the energy defined in Equation~\ref{eq:energy}. This generalization allows predictive-coding-based transformers to perform almost as well as standard transformers with the same model complexity \cite{pinchetti2022}. In computational neuroscience, the use of hybrid generative models (with discrete and continuous state spaces) is fairly established: see, e.g., \cite{friston2017graphical} for a first principles account, which can be regarded as a generalization of PC within the setting of well-defined hierarchical generative models with mixed discrete and continuous states. 

\begin{figure}[!t]
     \centering
     \begin{subfigure}[b]{0.25\textwidth}
         \centering
         \includegraphics[width=\textwidth]{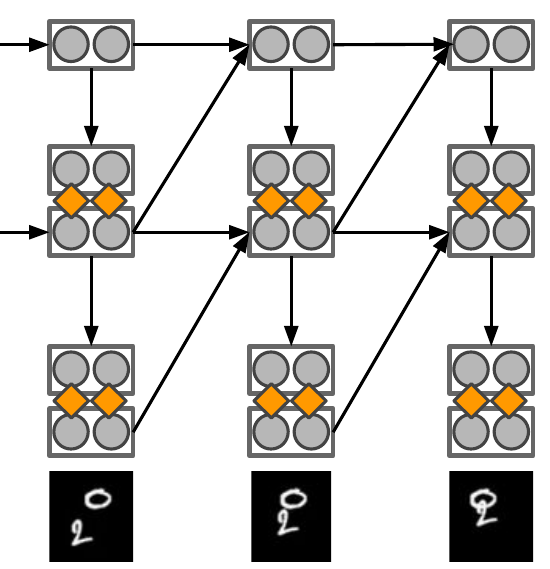}
         \caption{Temporal NGC.}
         \label{fig:tncn}
     \end{subfigure}
     \hfill
     \begin{subfigure}[b]{0.2\textwidth}
         \centering
         \includegraphics[width=\textwidth]{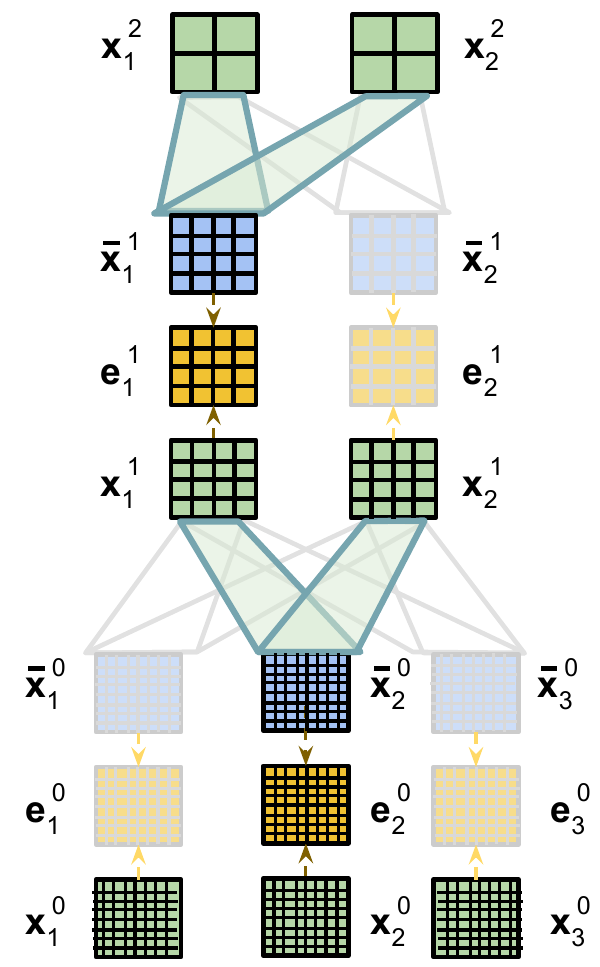}
         \caption{Convolutional NGC.}
         \label{fig:conv_ngc}
     \end{subfigure}
     \hfill
     \begin{subfigure}[b]{0.40\textwidth}
         \centering
     %     \includegraphics[width=\textwidth]{figs/sncn.pdf}
     %     \caption{Continual NGC.}
     %     \label{fig:sncn}
     % \end{subfigure}
     % \begin{subfigure}[b]{0.195\textwidth}
     %     \centering
     %     \includegraphics[width=\textwidth]{figs/actpc_agent.pdf}
     %     \caption{Active NGC.}
     %     \label{fig:actpc}
     % \end{subfigure}
     % \begin{subfigure}[b]{0.195\textwidth}
     %     \centering
         \includegraphics[width=\textwidth]{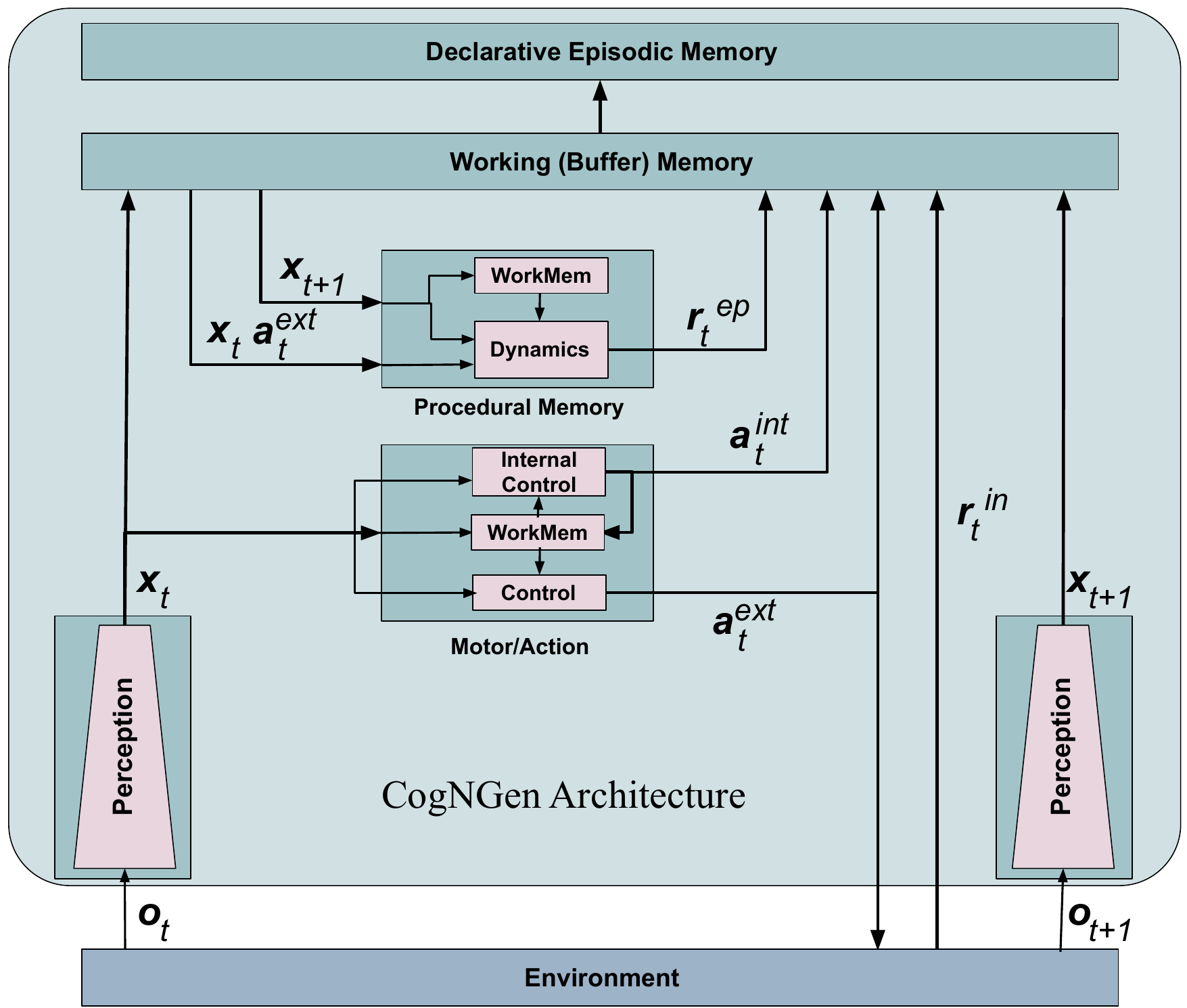}
         \caption{CogNGen.}
         \label{fig:cogngen}
     \end{subfigure}
     %\vspace{-0.15cm}
        \caption{Various systems based on NGC-based PC: (a) a temporal NGC model (orange diamonds depict error nodes, grey circles depict state nodes, while arrows represent synaptic projections) for processing temporally-varying sequences of data patterns, e.g., frames from the moving MNIST dataset, (b) a convolutional NGC model (figure adapted from \cite{ororbia2022convolutional}; green grid blocks represent state feature maps, blue grid blocks represent prediction maps, yellow-orange grid blocks represent error maps, and blue parallelograms depict deconvolutional pathways) for extracting distributed representations of natural images, 
        %(c) the sequential neural coding network (for continual learning, figure adapted from \cite{ororbia2019lifelong}), 
        %(d) active neural coding / ActPC (for robotic control, figure adapted from \cite{ororbia2022active}), 
        and (c) the CogNGen architecture (figure adapted from \cite{ororbia2022cogngen}), a modular control system that implements several core elements of the Common Model of Cognition \cite{ororbia2021towards} with PC and vector-symbolic memory ($\mathbf{r}$ depicts a reward,  while $\mathbf{a}$ depicts an action vector).}
        \label{fig:ngc_models}
\end{figure}

\paragraph{Computer Vision.} When processing images, PC has shown promise, particularly when generalizations that utilize convolutional operators are considered to tackle problems in vision such as object recognition and discrimination (using over-complete sparse representations) \cite{principe2014cognitive,chalasani2012temporal,sledge2021faster,chalasani2014context, spratling2017hierarchical}. Early NGC work used simplifying assumptions like greyscale images and low scene complexity with easily identifiable objects. NGC circuits can process image patches ($P \times P$ pixel grids) to learn low-level patterns such as edges and strokes. NGC was later generalized to natural images through convolutional neural generative coding (Conv-NGC), which integrates standard convolution/deconvolution operations \cite{ororbia2022convolutional}. In Conv-NGC, forward and feedback synapses manipulate feature maps coupled to error units that compute mismatch signals for iterative inference. Notably, Conv-NGC models learn feature representations that implicitly embody image pyramids \cite{adelson1984pyramid}.

\paragraph{Temporal Data.}
PC has achieved a discrete success in temporal sequence modeling. Early work developed hierarchical neural generative modeling for natural video data \cite{chalasani2013deep,chalasani2014context,santana2015parallel,chen2019minimum}, using top-down information to modulate lower-layer activities for locally invariant representations. This built on PC formulated as a Kalman filter \cite{rao1997dynamic,erdogmus2002modified} and deep hierarchies of local recurrent networks \cite{schmidhuber1991neural,schmidhuber1992learning}.
One of the earliest NGC formulations addressed time-varying data through the temporal neural coding network (TNCN) \cite{ororbia2017learning}, later generalized to the parallel temporal neural coding network (P-TNCN) \cite{ororbia2020continual}. Both models processed pattern sequences without temporal unfolding, making them local in space and time. They demonstrated zero-shot generalization to unseen sequences, with TNCN adapting to bouncing ball videos of different object quantities and P-TNCN synthesizing frames with varying quantities and qualities without altering synaptic weights. Recently, dynamic PC \cite{jiang2022dynamic} combined deep learning tools with mixture models, acquiring bio-plausible receptive fields for sequential visual perception tasks.

%So far, however, despite the theoretical basis as an extended Kalman filter and generalized filtering \cite{friston2010generalised}, most of the applications of PC in the literature are on static data, with .

\paragraph{Continual Learning.} Another promising aspect of  P-TNCNs described above is its ability to conduct continual sequential learning \cite{ororbia2020continual}, as its generative ability on previously seen sequence modeling tasks did not deteriorate nearly as much as in recurrent networks. Additional effort \cite{ororbia2019lifelong} explored strengthening the memory retention ability of NGC by examining the challenging problem of online cumulative learning, where datasets (or tasks) were presented to the system in the form of a stream and with no indication of when the task was switched. Surprisingly, the NGC-based model was shown to outperform or be competitive with a wide swath of backprop-based approaches that relied on memory-buffers (replay), regularization, and/or auxiliary generative modeling in order to preserve prior knowledge. Stronger performance was seen when another neural circuit drove the lateral recurrent synapses of S-NCN  based on competitive learning. This simple task mediating circuit was later improved in follow-up effort \cite{ororbia2021continual}. Another later effort developed a stochastic implementation of PC with readable/writable memory (BayesPCN) \cite{yoo2022bayespcn} that was shown to be robust to sample-level forgetting.
%Other variants of PC attempt to tackle the problem of continual learning by exploiting the probabilistic interpretation of its local prediction scheme, such as in BayesPCN (Bayesian PC networks) \cite{yoo2022bayespcn}.

\paragraph{Active Inference and Control.} In computational neuroscience, early formulations of active inference were based on equipping PC with reflexes in order to simulate a variety of behaviors; ranging from handwriting and its observation \cite{friston2011action}, through oculomotor control \cite{friston2012perceptions,perrinet2014active} to communication \cite{kiebel2009perception,isomura2019bayesian}, inspiring works that merge it with kinematic control \cite{priorelli2023deep,priorelli2023efficient}. Active NGC (ANGC) \cite{ororbia2022backprop} and active PC (ActPC) \cite{ororbia2022active} formulate active inference through predictive processing, contrasting with backprop-trained networks. ANGC implements modular agent design with policy and transition dynamics circuits learning from sparse rewards, while ActPC adds actor and prior preference circuits with working memory integration. Both work well on standard RL control and simulated robotics \cite{tani2016exploring,lanillos2018adaptive,oliver2021empirical}.
Parallel frameworks include active PC (APC) \cite{rao2022active} and active PC networks (APCNs) \cite{gklezakos2022active} for learning part-whole hierarchies from natural images and hierarchical reinforcement learning in multi-room grid-worlds. However, these utilize backprop in portions of their architectures. The COGnitive Neural GENerative system (CogNGen) \cite{ororbia2021towards,ororbia2022cogngen, ororbia2023maze} combines NGC with hyperdimensional memory models, implementing motor cortex, procedural memory, working memory buffers, and episodic memory modules, performing well on maze exploration tasks requiring cross-episode memory.

\begin{table}[!t]
    \centering
    \begin{tabular}{l | p{0.38\linewidth} | p{0.28\linewidth}} % 
    \toprule
    \textbf{Research Domain} & \textbf{Tasks} & \textbf{References}\\
    \midrule
    Discriminative Learning & 
        \begin{tabular}[t]{@{}l@{}}
             Categorization,\\
             Graph-based classification
        \end{tabular} 
    & \cite{whittington2017approximation,salvatori2022incremental,byiringiro2022robust} \\
    \midrule
    Natural Language Processing & 
        \begin{tabular}[t]{@{}l@{}}
             Language modeling,\\
             Masked language modeling
        \end{tabular} 
    & \cite{ororbia2020continual,pinchetti2022,araujo2021augmenting} \\
    \midrule
    Computer Vision & 
        \begin{tabular}[t]{@{}l@{}}
             Unsupervised reconstruction, \\
             Object classification, Feature extraction
        \end{tabular} 
    & \cite{schmidhuber1991neural,rao1999predictive,sledge2021faster,chalasani2014context,spratling2017hierarchical,principe2014cognitive,chalasani2012temporal,ororbia2022convolutional} \\
    \midrule
    Temporal Modeling & 
        \begin{tabular}[t]{@{}l@{}}
             Video sequence modeling,\\
             Sequence-based classification
        \end{tabular} 
    & 
        \begin{tabular}[t]{@{}l@{}}
             \cite{rao1997dynamic,erdogmus2002modified,chalasani2013deep,chalasani2014context,santana2015parallel} \\
             \cite{chen2019minimum,ororbia2017learning,ororbia2020continual,jiang2022dynamic}
        \end{tabular} 
         \\
    \midrule
    Lifelong Learning & 
        \begin{tabular}[t]{@{}l@{}}
             Continual classification, \\
             Continual sequence modeling
        \end{tabular} 
    & \cite{ororbia2019lifelong,ororbia2020continual,ororbia2021continual,yoo2022bayespcn} \\
    \midrule
    Control / Robotics &  
        \begin{tabular}[t]{@{}l@{}}
             Reinforcement learning, \\
             Robotic control, maze navigation
        \end{tabular} 
    & 
        \begin{tabular}[t]{@{}l@{}}
            \cite{lanillos2018adaptive,muhammad2015neural,li2020cognitive,ororbia2022backprop,gklezakos2022active,ororbia2021towards}\\
             \cite{oliver2021empirical,ororbia2022cogngen,ororbia2022active,rao2022active,ororbia2023maze}
        \end{tabular} 
         \\
    \bottomrule
    \end{tabular}
    \caption{Overview of PC utilized and developed for machine learning, grouped by application/topic area. For a comprehensive study of the state of the art results in computer vision, we refer to \cite{pinchetti2024benchmarking}.}
    \label{sec:research_topic_grouping}
\end{table}

\section{Plausibility of Predictive Coding in Neuroscience}
\label{sec:pc_neuro}

A common question in machine learning is: at what point can a particular algorithm be considered to be biologically plausible? This arises from the fact that no computer simulation can fully replicate the intricate workings of the brain in every respect, and, hence, there will invariably be certain nuances that render the simulation implausible in some way. Furthermore, different research agendas consider different properties for differentiating biologically plausible and non-plausible models. In this section, we will start by addressing the neuroscientific debate and problems with testing PC and then discuss the key properties of PC that satisfy this distinction and which do not.

\paragraph{Neuroscientific Debate over Predictive Coding.}
Despite the fact that PC is a subset of variational Bayesian methodology, there is far less direct experimental support for PC compared to the broader Bayesian brain hypothesis. While there is abundant evidence for the Bayesian brain hypothesis itself \cite{rao2002probabilistic,lee2003hierarchical,doya2007bayesian, seth2014cybernetic, friston2012history, clark2013whatever}, direct experimental support for PC across modalities remains more limited \cite{gabhart2025predictive}.
Neuroimaging does support some elements of PC \cite{gordon2017neural,walsh2020evaluating,caucheteux2023evidence}, particularly through the framing of mismatch negativity \cite{garrido2009mismatch,wacongne2012neuronal}. However, mismatch negativity can also be explained via neural adaptation without invoking hierarchical Bayesian mechanics \cite{gabhart2025predictive}. Evidence has emerged indicating that PC might be better considered as a cognitive, higher-level explanation rather than a lower-level sensory-based one \cite{gabhart2025predictive}.
Although PC can account for various neurophysiological phenomena such as bistable perception \cite{weilnhammer2017predictive}, perceptual illusions \cite{watanabe2018illusory}, and properties of V1 neuronal responses \cite{spratling2010predictive}, fewer studies explicitly test hypotheses inherent to PC. These core hypotheses include expectation-scaled error-signaling neural responses and top-down signaling representing sensory prediction.
PC has been criticized as difficult to falsify \cite{kogo2015predictive}. 

The translation of PC from algorithmic specification to biophysical instantiation is often unclear, creating further ambiguity about which implementation is being tested in particular studies. Some PC variants better explain certain neurophysiological data over others \cite{spratling2019fitting}, making it difficult for neuroscience researchers to test its validity \cite{walsh2020evaluating}.
Some studies suggest that PC is more likely to occur in certain brain regions depending on the sensory modalities being processed \cite{ficco2021disentangling}. More importantly, there appears to be no evidence of a distinction between predictions and errors at the network level \cite{ficco2021disentangling,heilbron2018great}.
Nevertheless, despite this unsettled debate surrounding PC and its empirical grounding, advances in neurophysiological methodology offer promise. Higher resolution fMRI, calcium imaging, and other techniques, when paired with anatomical models and simulation, might enable future derivations of finer-grained hypotheses. These advances could also facilitate comparisons across variant PC implementations that share the same underlying theoretical framework, including those with different forms of encoding or representation of prediction errors.

\paragraph{Error Neurons.} Our understanding of how PC might be implemented at the neuronal level has certainly changed over the last decade. The initial assumption was that the brain would encode two families of neurons/structures: one in charge of propagating predictions and one in charge of propagating errors \cite{rao1999predictive, whittington2017approximation}. As of today, we have no definitive empirical evidence of the existence of single error neurons, though there is ample evidence for laminar-specific segregation of neuronal populations that may communicate predictions and errors, respectively \cite{bastos2012canonical,shipp2016neural}. %In addition, one could read the activity of midbrain dopamine neurons as reporting a certain kind of prediction error \cite{schultz2000neuronal,bayer2005midbrain}. 
Recent work has shown that error signals could potentially be computed by the local voltage dynamics in the dendrites \cite{schiess2016somato,sacramento2018dendritic,mikulasch2023error,bastos2020layer}. For a more detailed description in the context of a possible biological neural implementation, see \cite{whittington2019theories, mikulasch2023error}. Although existing technologies make it difficult to empirically demonstrate PC at the level of a single neuron, we have stronger evidence of PC at the level of neural populations, or brain regions \cite{walsh2020evaluating,gabhart2025predictive}. PC can account for multiple brain phenomena, such as end-stopping and extra-classical receptive fields effects in V1 \cite{rao1999predictive}, bistable perception \cite{hohwy2008predictive}, illusory motion \cite{watanabe2018illusory}, repetition-suppression \cite{auksztulewicz2016repetition}, and attentional modulation of cortical processing \cite{feldman2010attention,kanai2015cerebral}. It has also been shown that, when listening to speech, the human brain stores acquired information in a hierarchical fashion, where frontoparietal cortices predict the activities of higher level representations \cite{caucheteux2023evidence}. 

\paragraph{Precision Engineering.} Another challenge in the implementation of PC is the management and updating of precision weighting. Typically, the covariance matrix $\mathbf{\Sigma}^\ell$ or its inverse, the precision matrix $(\mathbf{\Sigma}^\ell)^{-1}$, must be adjusted or computed using matrix inversion \cite{rao1997dynamic,ororbia20,tang2023recurrent}. However, implementations of PC in computational neuroscience have solved this problem using standard solutions from precision or covariance component analysis in a biologically plausible way: see \cite{friston2008hierarchical} (Equation 57) and \cite{bogacz2017tutorial,millidge2021predictive} for details. This is an important aspect of PC, because precision weighting is thought to implement attention in a neuroscience setting. There is a substantial literature on PC in this setting that might usefully inform machine learning implementations \cite{feldman2010attention,brown2013active,moran2013free,kanai2015cerebral,parr2019attention,barron2020prediction}.

\paragraph{Synaptic Constraints.} In standard PC networks, synaptic values, which are crucial for neural computation, are generally not subject to strict limitations after an update. This can cause the values to become highly positive or negative, which makes the model less stable overall.  This is tackled in NGC, which introduces a constraint to ensure that the Euclidean norm of any row or column in the synaptic matrix does not exceed one, aiding in stability, a practice rooted in early classical sparse coding linear generative models \cite{olshausen1997sparse}. Beyond this, there have been recent attempts to establish heuristic limits on the magnitude of synaptic values in order to thwart declines in classification performance \cite{kinghorn2023preventing}. A second implausibility of PC models is the frequent alterations in the signs of their synapses (or ``sign-flipping''), that can go from negative to positive (and vice versa) during training, an aspect that is pivotal in emulating real cortical functionality. A potential solution to prevent sign-flipping would be to enforce non-negativity constraints on synaptic values and explicitly model groups of excitatory and inhibitory neurons, given that this configuration would apply the necessary positive and negative forces essential for PC inference and learning \cite{ororbia2019spiking,alonso2021tightening,ororbia2023learning,cornford2024brain}.

\section{Software Frameworks and Novel Hardware}
\label{sec:software_hardware}

Software and hardware play crucial roles in enabling innovation and practical implementation of computational intelligence models. While frameworks like PyTorch \cite{torch} and TensorFlow \cite{tensorflow} have been instrumental for deep learning, PC research faces unique challenges due to sparse software support and evolving hardware opportunities. In terms of software, available frameworks remain relatively sparse, with most research resulting in scattered, paper-specific code that hinders wider adoption and reproducibility. Nevertheless, several libraries aim to democratize PC research:  \textbf{\textit{ngc-learn}} \cite{ngclearn} is the official library for NGC, is grounded in neuronal cable theory and supports arbitrary PC model and neuromorphic system construction as well as general neural circuit simulation. \textbf{\textit{PCX}} \cite{pinchetti2024benchmarking} is a deep learning oriented library, is built on JAX/Equinox and enables plug-and-play deep learning with PC updates and demonstrates state-of-the-art results across frameworks. Other libraries include \textbf{\textit{pypc}} \cite{pypc} for hierarchical PC models, \textbf{\textit{predify}} \cite{predify} for converting deep architectures to PC-like systems, and \textbf{\textit{pyhgf}} \cite{legrand2024pyhgf} for stochastic and Bayesian PC formulations, and the recently proposed jPC \cite{innocenti2024jpc} for fully parallel deep learning architectures.

\begin{figure}[!t]
\centering
\includegraphics[width=5.35in]{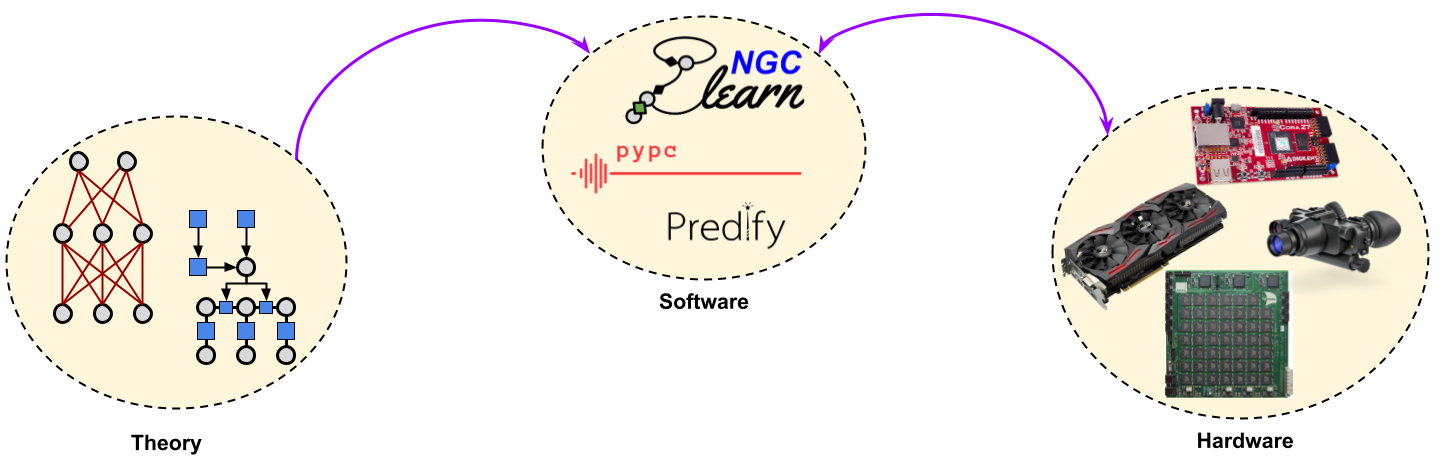}
\caption{The theory-software-hardware cycle of predictive processing showing bidirectional relationships between theoretical development, simulation software, and hardware platforms.}
\label{fig:pc_software_cycle}
\end{figure}

\subsection{Novel Hardware Implementation}

Hardware limitations historically shape research directions \cite{lecun19hardware}. While GPUs and TPUs dominate current training, emerging technologies like memristors, spintronics, and optics could revolutionize the field \cite{memristor,strukov2008missing,grollier2020neuromorphic}. PC is particularly well-suited for alternative hardware as an energy-based model trainable through equilibrium propagation \cite{scellier17, zucchet2022beyond}. Its iterative inference with layer-wise parallel computation, Hebbian adaptation, and local energy optimization enables significant parallelization, reducing communication bottlenecks inherent in backpropagation. This alignment with neuromorphic hardware creates opportunities for energy-efficient learning \cite{kendall2020}, with dynamical memristors offering promising potential for brain-inspired systems \cite{kumar2022dynamical}. Beyond conventional hardware, PC could enable "intelligence-in-a-dish" technology through adaptive computation in biophysical mediums \cite{haselager2020breeding,smirnova2022neuronal,smirnova2023organoid}. Given the association of PC  with cortical computation \cite{spratling2010predictive,bastos2012canonical,shipp2013reflections}, it is inherently compatible with organoid intelligence \cite{friston2023sentient}. Early spike-level implementations \cite{ballard2000single,ororbia2019spiking} show promise for dynamic predictive processing. Organoid growth and decay could facilitate model selection \cite{wasserman2000bayesian}, potentially improving generalization through natural evolution of neural structures.

\begin{figure}[!t]
\centering
\includegraphics[width=5.2in]{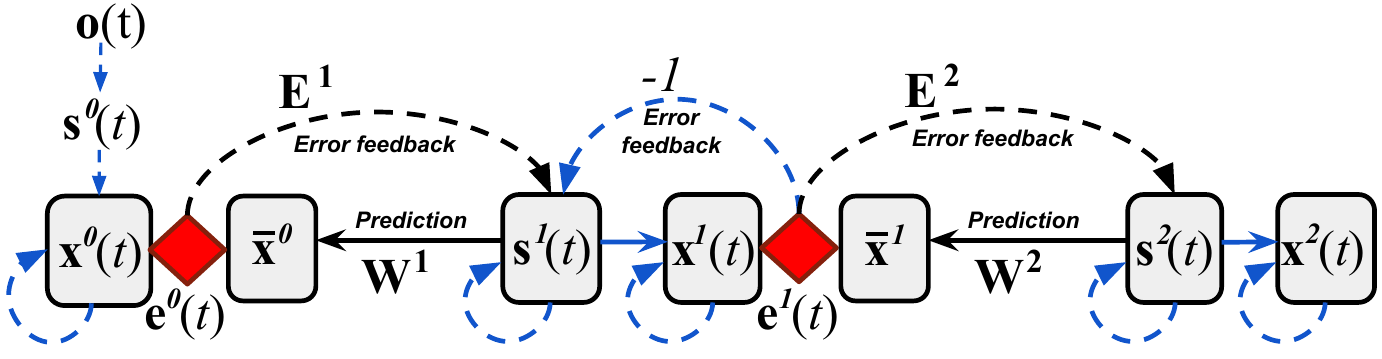}
\caption{Spiking neural coding formulation where $\mathbf{s}^\ell(t)$ represents spike nodes and $\mathbf{x}^\ell(t)$ are trace approximations. Solid arrows show forward synapses, dashed arrows show error transmission.}
\label{fig:spncn}
\end{figure}

\paragraph{Spiking Predictive Coding.} A key challenge involves generalizing PC to spike-level processing. Biological neurons communicate through sparse action potentials creating information-rich spike trains \cite{de2007correlation}, inspiring spiking neural networks that encode information through precise timing \cite{maass1997networks}. This enables energy-efficient neuromorphic hardware \cite{furber2014spinnaker,merolla2014truenorth,davies2018loihi}, unlike energy-intensive GPU-based networks. However, most PC formulations fail to account for sparse, discrete communication, reflecting the nature of PC as describing ensemble rather than individual neuron dynamics \cite{lee2003hierarchical}. The spiking neural coding framework attempts PC implementation in spiking networks using spike-induced synaptic adjustments \cite{ballard1999model,ororbia2019spiking}. Unlike standard spike-timing-dependent plasticity requiring hand-crafted layers \cite{bi1998synaptic}, this framework uses flexible synaptic conductance models, membrane potentials, and trace mechanisms. Each layer estimates the trace activity of the following layer, passing discrepancies through evolved error feedback synapses (Fig.~\ref{fig:spncn}). The framework adapts to various spiking functions from leaky integrators to Hodgkin-Huxley models \cite{hodgkin1952quantitative}. Recent implementations \cite{lan2022pc,mikulasch2022dendritic,mikulasch2023error} operate in discriminative contexts but do not address weight transport, sharing feedback with forward synapses. For comprehensive coverage, see \cite{n2024predictive}.

\section{Looking Forward: Important Directions for Predictive Coding Research}
\label{future}

As we have discussed in Sections~\ref{sec:pc_frameworks} and \ref{sec:pc_applications}, the performance of PC models can be quite below that of modern-day deep neural networks trained with backprop. The future of PC in machine learning strongly depends on our ability to address and fill this gap. Specifically, among the main directions for future work related to PC should be understanding the fundamental reasons behind this performance mismatch and using this acquired insight to develop and design novel PC schemas, mathematical frameworks, and computational models that work in large-scale settings, where current deep learning models excel. 

\paragraph{Efficiency.} The first drawback of PC is its efficiency. This is a consequence of its underlying iterative inference process, which generally needs to be run until convergence. In practice, it is common to let a PC model run for a fixed number of iterations $T$, but this number must be large in order to reach high performance, and deeper networks require more iterations in order to perform well. To this end, it would be useful to derive different optimization techniques and methodologies that carry out the minimization of the variational free energy. Such techniques could come in the form of (faster) variations of the expectation-maximization (EM) algorithm (equivalent to gradient descent in the standard case). An example of such a scheme is the newly proposed \emph{incremental} PC \cite{salvatori2022incremental}, based on incremental EM. The convergence of this method has also been proven using methods from dynamical systems, under the assumption of sufficiently small parameters \cite{frieder2022non}. Future work could investigate whether there are better alternatives or whether the optimal update rule can be learned with respect to specific tasks and datasets (drawing inspiration from the recent developments in deep meta-learning \cite{vilalta2002perspective,santoro2016meta,hospedales2021meta}). More advanced amortized inference algorithms and mechanisms might prove to be another critical ingredient for reducing the computational burden of the PC inference process itself, as, historically, the iterative inference of classical sparse coding linear generative models \cite{olshausen1997sparse} was dramatically sped up when introducing a second recognition neural model \cite{kavukcuoglu2010fast}. It might be that the answer may lie in some variations of the EM algorithm that are lesser known to the machine learning community, such as \emph{dynamical} expectation maximization (DEM) \cite{anil2021dynamic,friston2008variational,friston2008variational}, or in alternative, more efficient implementations of precision-weighted prediction errors, as in \cite{jiang2022dynamic}. Other possibilities may lie in methods based on different message passing frameworks, such as belief propagation and variational message passing over factor graphs.

\paragraph{Optimization Tricks and Heuristics.}
Future research will also need to focus on the study of optimization techniques that have proven useful and invaluable for variational inference, such as those that promote the inclusion of precision weighting parameters into the picture \cite{friston2012predictive}.
In the same spirit, recent progress on natural-gradient methods for gradient-based variational inference—ranging from stochastic NGVI with non-asymptotic guarantees, through Kronecker-factored curvature approximations, to the variational predictive natural gradient—offers curvature-aware updates that can be interpreted as learning layer-wise precisions online, and therefore provides an immediately transferable optimisation tool-set for scaling precision-weighted predictive coding \cite{khan2018fast,osawa2020scalable,tang2019variational}.
Despite interesting developments obtained over the last several years, and despite the fact that precisions are of primary importance in simulations used in the neurosciences, these techniques have only been tested in small and medium scale settings \cite{ofner2021predprop,alonso2023understanding,zahid2023curvature}.
More generally, the field of deep learning has enjoyed immense benefits from simple optimization tricks developed throughout the last decade, such as dropout \cite{srivastava2014dropout}, batch normalization \cite{ioffe2015batch}, adaptive learning rates such as Adam \cite{kingma2014adam} and RMSprop \cite{tieleman2012lecture}, and the introduction of the ReLU activation \cite{fukushima1969visual} and its variants.
Without these techniques, emerging from the joint effort of thousands of researchers, the training of extremely overparametrized neural models would not lead to the results that we observe today.
Interestingly, backpropagation itself exhibited many flaws and limitations at the beginning that precluded it scaling to high-dimensional data spaces: in the late 90s, kernel learning methods were dominant and generally found to be more effective than artificial neural networks, which were generally disregarded due the problem of vanishing gradients \cite{pascanu2013difficulty} as well as their strong requirement for heavy computational power.
Most of these problems have now been addressed in over thirty years of research, again through a collective effort of thousands of people.
Given the history of deep learning, one might ask: what are the dropout, batch norm, and Adam optimizer equivalents for PC?
Addressing such a question would be of vital importance if the goal is to scale the applicability of PC, and will hopefully become a more prominent topic of interest of more research efforts going forward.

\paragraph{Stochastic Generative Models and Sampling.} A different but important, direction, is that related to generative models. We have been discussing the importance of putting statistical models of joint distributions back into the research agenda of generative AI; however, most of the research efforts that go beyond small-scale tasks  focus on supervised learning \cite{salvatori2022incremental, pinchetti2022} (with small notable exceptions, such as the unsupervised neural circuit of \cite{ororbia2022convolutional}). To do so, we need to develop models that are capable of sampling data points from a well-computed posterior distribution, using simulation methods such those based on Langevin dynamics \cite{song2019generative}. This would benefit different subfields of Bayesian inference, such as out-of-distribution (OOD)  detection, uncertainty minimization, and data reconstruction. In fact, PC is particularly suitable to OOD detection, thanks to a measure of surprise that is always readily available to the model, that is, with respect to its variational free energy. 

Another probabilistic generative modeling research direction for PC is to incorporate the uncertainty about model (i.e., synaptic) parameters into the variational bound on marginal likelihood. That is, one can equip synaptic parameters with probability distributions (as opposed to using point estimates) \cite{tschantz2025bayesian}. This move would take PC closer to its Bayesian roots: current machine learning implementations of PC do not treat the model parameters as random variables and can therefore be regarded as an expectation maximization (EM) procedure, where the M-step ignores uncertainty about the parameters \cite{dauwels2007variational}. The benefit of treating parameters as random variables is that one can evaluate the model evidence required for structure learning \cite{tenenbaum2011grow,tervo2016toward,smith2020active,rutar2022structure}, i.e., one could place prediction errors on the synaptic weights and evaluate the resulting variational free energy (or marginal likelihood) for structure learning or, in a biological setting, morphogenesis \cite{kuchling2020morpho,kiebel2011free}. This research direction takes PC into the world of generalized filtering and dynamic causal modeling, in which parameters can be regarded as latent states that change very slowly \cite{friston2010generalised,stephan2010ten}. In the case of dynamic causal modeling (i.e., the variational inversion of state-space models under Gaussian assumptions), structure learning can be rendered particularly efficient through particular instances of Bayesian model selection, which provides a principled basis for comparing alternative generative hypotheses \cite{hoeting1999bayesian,gershman2012tutorial,penny2012comparing}.  
In predictive coding (PC), the same logic appears as \emph{Bayesian model reduction}, where synaptic weights that receive little posterior evidence are pruned from a full (parent) network \cite{friston2011post}.  
Looking ahead, a natural extension is to replace these fixed-dimension reductions with \textbf{Bayesian non-parametric} (BNP) priors—such as Dirichlet or Indian-buffet processes—that equip each PC layer with an \emph{unbounded} supply of latent causes or synaptic motifs \cite{gershman2012tutorial,yeh2006dirichlet,goldwater2007nonparametric}.  
Thus in BNP–PC hybrid, precision-weighted prediction errors might simultaneously (i) adjust beliefs about existing components and (ii) signal when residual error warrants “breaking a new stick’’ (creating a component) or dropping a redundant one.  
Thus, BNP priors transform PC into a self-organising system that learns both its parameters \emph{and} its effective dimensionality online, unifying synaptic growth and pruning within a single, neurally plausible free-energy minimisation scheme.

% Bayesian model selection \cite{hoeting1999bayesian,gershman2012tutorial,penny2012comparing}; namely, Bayesian model reduction to remove or prune redundant parameters from a full or parent model \cite{friston2011post}. Related generalizations of PC might also leverage advances in nonparametric Bayes to learn the structure of neuronal networks. Please see \cite{gershman2012tutorial,yeh2006dirichlet,goldwater2007nonparametric} for further details.

\paragraph{Crafting Control Systems.} PC brings with it the promise of learning a powerful generative model that is continuously and iteratively refined as more sensory samples are gathered over time. This has led some early work to consider such a process as the basis for world models that drive modular, brain-motivated cognitive models, capable of combining perception and action in the context of playing video games \cite{principe2014cognitive,li2020cognitive} and robotic control tasks \cite{tani2016exploring,ororbia2022backprop,ororbia2022active}, as well as large-scale cognitive architectures \cite{laird2017standard,ororbia2022cogngen,ororbia2023maze}. This has important implications beyond machine learning, particularly for the domains of cognitive science and cognitive neuroscience where a key pathway is to craft computational theories of mind and to examine their fit to human subject data on controlled psychological tasks \cite{schwartenbeck2016computational} as well as their ability to generalize to cognitive functionality. Hence, a promising future direction is to design modular, increasingly complex systems made up of PC circuitry, that rely on single fundamental PC circuits and low-level dynamics. 

There are important lessons that come from crafting systems of PC circuitry that can inform the fundamentals of PC itself. For example, in the PC-centric cognitive architecture of \cite{ororbia2022cogngen}, it was found that the synergy of PC with another important neural model, i.e., vector symbolic memory \cite{kanerva1988sparse,levy2008vector,kanerva2009hyperdimensional}, facilitated effective complex auto-associative and hetero-associative memory operations that led to rapid convergence on complex maze navigation tasks and, furthermore, facilitated the design of a novel PC circuit that leveraged its neural dynamics to learn a policy for internally manipulating a flexible recurrent memory system \cite{ororbia2023maze}. 

Using PC as a fundamental neural building block for control systems is still in its earliest stages, and although more effort will be needed for it to become viable for building robust computational theories of mind, it presents a worthwhile long-term direction for crafting simulated natural intelligence. This promise is made even stronger when considering that deep backprop-based networks are beginning to find use in the field of cognitive science \cite{ma2020neural}. Making developments along this direction would also crucially benefit the research in active inference \cite{friston2016active,friston2017graphical,da2022active,friston2022designing}, given that agent systems that engage in epistemic foraging often center around the use of dynamic generative models and can even be viewed as simple control models.% - this would motivate further effort in scaling up the fundamental learning and inference dynamics of PC in order to support the more challenging forms of active inference, such as those that learn from raw, unstructured data in the context of partially observable Markov decision processes.

\section{Conclusion and Outlook}
\label{sec:conclusion}

%Why does this survey exist? PC is very promising: despite the fact that BP is amazing, there could be alternatives, such as BP. Plus, PC is grounded in theory: it follows all the models developed in statistics in the 60s. Can we use these methods large scale? BP does not, but maybe it could be nice.
%Summary of general picture: from generative model to neuroscience. What’s next?
Generative models will most likely play a large role in the future of artificial intelligence. Despite this, current research seems to mostly focus on a limited class of models, overlooking possible alternatives. On the one hand, this is justified by the exceptional results obtained by deep, backprop-trained artificial neural networks; on the other hand, more principled methods, whose formulations are grounded in statistics and information theory, may deserve attention. In this work, we have focused on inverting a specific class of continuous-state generative models, namely, predictive coding. More precisely, we have: 
(1) presented a review of predictive coding schemes in machine learning, highlighting previous art that has led us to the point that we are at today, 
(2) summarized important open questions in the field that need to be answered in order to unlock and utilize the full potential of predictive coding, and 
(3) identified possible applications and future research directions. As of today, we are seeing some of the potential of PC, but we are still far away from large-scale applications that may call for a significant investment in PC research. Already in the past years, we have seen impressive growth in the field of predictive coding in machine learning: it was only in $2017$ that Whittington and Bogacz demonstrated how to train a small predictive coding classifier on MNIST \cite{whittington2017approximation}. In addition, we have already seen how integrating a few ideas from predictive coding can lead to interesting and more powerful backprop-based deep models \cite{lotter2016deep,sato2018short,han2018deep,zhong2018afa,rane2020prednet,araujo2021augmenting}. 

%What ‘tricks’ are we missing? What is the layernorm, or dropout of PC, that make these models learn well, and scale up?

%Encouraging other researcher to work on open problems (motivation!). PC is very promising, and we need more effort because it does not matter how good a method is, without a good research community we do nothing. 
%More effort in development: library to do machine learning;
One of the main goals of this survey is to encourage researchers to build on the results of decades of prior effort and focus on the challenges offered by predictive coding: no success story can come without an active community behind it, as evidenced by backpropagation-centric deep learning. We remark that a promising methodology is only as good as the efforts that are made to advance it, both empirically and theoretically. Community effort will be needed to advance predictive coding from both the software and the hardware standpoint; particularly, to develop computational schemes that exploit the advantages on offer, such as its parallelism and sparse, local, and potentially energy-efficient computations. Although research progress in predictive coding has been consistent over the past decades, we may be only starting to realize the benefits afforded to artificial intelligence by reverse engineering the cortex and other biological structures.
% Summary on the versions of PC proposed: why are they different? Why are they similar?
% Computational intelligence: improving AI by reverse engineering the cortex.
%Although research progress in predictive coding has been consistently over the past 30+ years, resulting in many task-level models, machine learning achievements, and the three general computational frameworks we have surveyed in this work, we remark that we are only but observing the start of the promise afforded by improving artificial intelligence and developing robust computational intelligence by reverse engineering the cortex and other biological neural structures.

\section*{Acknowledgments}
Thomas Lukasiewicz
was supported by %the Alan Turing Institute under
%the UK EPSRC grant EP/N510129/1, 
the AXA Research Fund. %, and the EU TAILOR grant 952215. We further acknowledge the support of the Cisco Research Gift Award \#26224.

% conclusion notes
\begin{comment} 
Conclusion:
Summary on the versions of PC proposed: why are they different? Why are they similar?
Summary of general picture: from generative model to neuroscience. What’s next?
Encouraging other researcher to work on open problems (motivation!). PC is very promising, and we need more effort because it does not matter how good a method is, without a good research community we do nothing. 
More effort in development: library to do machine learning;
What ‘tricks’ are we missing? What is the layernorm, or dropout of PC, that make these models learn well, and scale up?
Why is this survey existing? PC is very promising: despite the fact that BP is amazing, there could be alternatives, such as BP. Plus, PC is grounded in theory: it follows all the models developed in statistics in the 60s. Can we use these methods large scale? BP does not, but maybe it could be nice.
Computational intelligence: improving AI by reverse engineering the cortex.
\end{comment}

\bibliographystyle{customunsrt}  
\bibliography{ref}

%\appendix
%\newpage
%\input{appendix_core}

\end{document}